\documentclass[lettersize,journal]{IEEEtran}
\usepackage{amsmath,amsfonts}
\usepackage{algorithmic}
\usepackage{algorithm}
\usepackage{array}
\usepackage[caption=false,font=normalsize,labelfont=sf,textfont=sf]{subfig}
\usepackage{textcomp}
\usepackage{stfloats}
\usepackage{url}
\usepackage{verbatim}
\usepackage{graphicx}
\usepackage{cite}
\usepackage[capitalize]{cleveref}
\usepackage{xspace}
\usepackage{paralist, tabularx}
\usepackage{multirow}
\usepackage{booktabs}
\usepackage{color}
\usepackage{placeins}
\usepackage{bbm}
\usepackage{tikz}
\makeatletter
\DeclareRobustCommand\onedot{\futurelet\@let@token\@onedot}
\def\@onedot{\ifx\@let@token.\else.\null\fi\xspace}

\def\eg{\emph{e.g}\onedot} 
\def\ie{\emph{i.e}\onedot} 
 
\def\etc{\emph{etc}\onedot} \def\vs{\emph{vs}\onedot}

\makeatother

\newcommand{\miaojing}[1]{{#1}}

\begin{document}
\title{Boosting Zero-shot Learning via Contrastive Optimization of Attribute Representations}

\author{Yu Du,
        Miaojing Shi$^\dagger$, ~\IEEEmembership{Senior Member,~IEEE,}
        Fangyun Wei, 
        and~Guoqi Li
\thanks{$^\dagger$Corresponding author.}
\thanks{Y. Du is with Tsinghua University. E-mail: duyu20@mails.tsinghua.edu.cn.}
\thanks{M. Shi is with the College of Electronic and Information Engineering, Tongji University, China. E-mail: miaojing.shi@kcl.ac.uk.}
\thanks{F. Wei is with Microsoft Research Asia. E-mail: fawe@microsoft.com.}
\thanks{G. Li is with Institute of Automation, Chinese Academy of Sciences (CAS), E-mail: guoqi.li@ia.ac.cn.}
}

\markboth{IEEE Transactions on Neural Networks and Learning Systems}%
{Du \MakeLowercase{\textit{et al.}}}


\maketitle

\begin{abstract}
Zero-shot learning (ZSL) aims to recognize classes that do not have samples in the training set. One representative solution is to directly learn an embedding function associating visual features with corresponding class semantics for recognizing new classes. Many methods extend upon this solution, and recent ones are especially keen on extracting rich features from images, \eg attribute features. These attribute features are 
normally extracted within each individual image; however, the common traits for features across images yet belonging to the same attribute are not emphasized. In this paper, we propose a new framework to boost ZSL by explicitly learning attribute prototypes beyond images 
and contrastively optimizing them with attribute-level features within images. Besides the novel architecture, two elements are highlighted for attribute representations: a new prototype generation module is designed to generate attribute prototypes from attribute semantics; a hard example-based contrastive optimization scheme is introduced to reinforce attribute-level features in the embedding space. 
We explore two alternative backbones, CNN-based and transformer-based, to build our framework and
conduct experiments on three standard benchmarks, CUB, SUN and AwA2. Results on these benchmarks demonstrate that our method improves the state of the art by a considerable margin. Our codes will be available at \textcolor{magenta}{ \emph{\url{https://github.com/dyabel/CoAR-ZSL.git}}}.
\end{abstract}


\section{Introduction}
\label{sec:introduction}
\IEEEPARstart{V}{isual} 
recognition flourishes in the presence of deep neural networks (DNNs)~\cite{krizhevsky2012imagenet,simonyan2014very,he2016deep}. Great success has been achieved with the availability of large amount of data and efficient machine learning techniques. The learned models are good experts to recognize visual classes that they were trained with, nonetheless, often fail to generalize on unseen classes with few or no training samples.  To tackle this, several learning paradigms such as few-shot learning~\cite{snell2017prototypical,yang2020restoring,sung2018learning}, zero-shot learning~\cite{romera2015embarrassingly,xian2017zero,wang2019survey}, self-supervised learning~\cite{he2020momentum,chen2021exploring} are introduced given different data/label availability during training and testing. This paper focuses on zero-shot learning (ZSL) where models are learned with sufficient data of seen classes and deployed to recognize unseen classes. An extended setting to ZSL is referred to as the generalized zero-shot learning (GZSL) where learned models are required to recognize both seen and unseen classes at testing stage. GZSL is closer to the recognition problem we face in the real world. Below, unless specified, we use ZSL to refer to both settings.

\begin{figure}[t]
\centering
\includegraphics[width=0.9\linewidth]{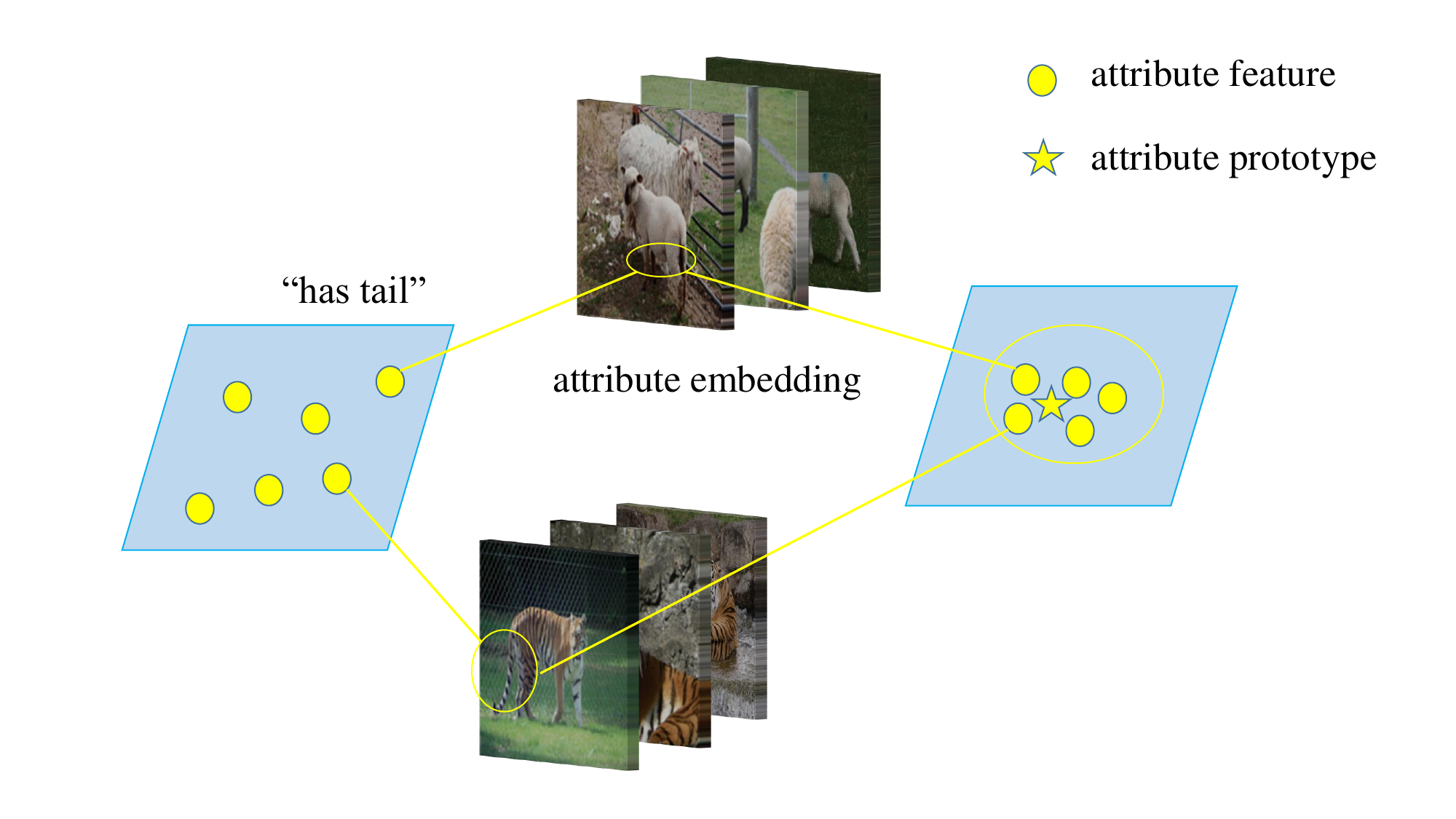}
  \caption{In previous methods, for different images, their attribute features corresponding to the same attribute (\eg \emph{has tail}) may be scattered in the embedding space, making them less representative for the attribute (left side). We solve this problem by explicitly learning  attribute prototypes beyond images and attribute-level features within images. They are jointly optimized to improve the attribute representations.  
}
\label{fig:intro}
\end{figure}

Despite that no training samples are provided for unseen classes in ZSL, the semantic information is available both sides to relate seen and unseen classes. The essential idea is to transfer knowledge from seen classes to unseen classes which can be through either the visual-to-semantic mapping~\cite{xu2020attribute,liu2021goal,yu2017transductive} or reversely the semantic-to-visual mapping~\cite{skorokhodov2020class,zhang2017learning,chen2021hsva}. Commonly used semantic information includes human-defined attributes of classes~\cite{lampert2009learning} and machine-generated word vectors of classes~\cite{church2017word2vec}.

A number of representative methods in ZSL make use of generative models to synthesize visual features for unseen classes based on their class semantics~\cite{vyas2020leveraging,xian2018feature,chen2021semantics}. Generative models can be hard to optimize as there exists a substantial gap between virtual features and features from real images. 
To bypass the use of generative models, other works tend to directly learn an embedding function associating visual features with corresponding semantic features~\cite{xu2020attribute,xie2020region,liu2020attribute,liu2021goal,chen2022msdn}. For instance, ~\cite{xu2020attribute,liu2021goal} proposed to encode each image (object) into a number of visual attribute features and then align them with class semantics. Because of the enriched representations of visual objects, these methods have reported state of the art results. {Despite their success, they still face challenges: attribute features are extracted within each individual image; due to the intra-attribute variance, one attribute might appear in form of multiple scattered features in the embedding space (see \cref{fig:intro}). This weakens the feature representativeness, hence impairs the classification. To tackle it, we propose to boost zero-shot learning via contrastive optimization between attribute prototypes across images and attribute-level features within images. As a result, the knowledge of attributes can be robustly learned and transferred from seen classes to unseen class in ZSL. }

Our paper follows embedding-based methods~\cite{xu2020attribute,xie2020region,liu2020attribute,liu2021goal,chen2022msdn} in ZSL. The essential motivation is to en-power the representations of visual objects such that those from different classes or with different attributes can be easily discriminated, while those from the same class or with same attributes can be easily connected. To achieve this, we propose a new framework, CoAR-ZSL, to explicitly learn attribute prototypes and attribute-level features separately and boost ZSL via the \textbf{C}ontrastive \textbf{o}ptimization of these \textbf{A}ttribute \textbf{R}epresentations. Specifically, a prototype generation module is firstly designed to take inputs of semantic features for attributes and generate corresponding attribute prototypes. Meanwhile, an input image is passed through a DNN to extract attribute-level visual features via an attention-based attribute localization scheme inspired by~\cite{xie2019attentive,xie2020region}. Attribute-level features are optimized against corresponding attribute prototypes using a contrastive triplet loss. To further reinforce these features, another hard example-based contrastive loss is introduced to optimize them across images in the embedding space. Besides the optimization of attribute representations, we also follow~\cite{jiang2018learning,zhang2017learning} to extract class-level features from images and optimize them with corresponding class prototypes; the latter are obtained jointly with attribute prototypes through the proposed prototype generation module.
We explore two alternative backbones for extracting class- and attribute-level features: the CNN-based and the transformer-based. Both versions outperform the state of the art in the respective ZSL and GZSL setting.

To summarize, the contribution of this paper is three-fold:
\begin{compactitem}
   \item Principally, we design a new framework to boost ZSL by explicitly learning attribute prototypes \emph{beyond} images and contrastively optimizing them with attribute-level features \emph{within} images. The concept of attribute prototype is distinguished from that of attribute-level feature for the first time. 
   \item Two elements are highlighted for attribute representations: a prototype generation module and a hard example-based contrastive optimization scheme. They are responsible for generating attribute (and class) prototypes and reinforcing attribute-level features, respectively.  
  \item We build our framework upon two alternative backbones, CNN-based and transformer-based, and evaluate on several standard benchmarks, CUB~\cite{welinder2010caltech}, SUN~\cite{patterson2014sun} and AwA2~\cite{xian2018zero}. Both versions of our method outperform the state of the art by considerable margins in the ZSL and GZSL settings.      

\end{compactitem}
\section{Related Works} \label{sec:relatedwork}
\subsection{Zero-shot learning} 
Modern ZSL methods can be broadly categorized as either generative-based or embedding-based~\cite{pourpanah2020review}. Many generative-based methods~\cite{vyas2020leveraging,li2019leveraging,xian2018feature,kong2022compactness} synthesize visual features of {unseen classes via the visual features of seen classes and the semantics of both seen and unseen classes}. There are also generative-based methods that synthesize features for virtual classes using strategies like MixUp~\cite{elhoseiny2021imaginative,elhoseiny2021cizsl++} such that prior unseen class semantics are no longer required to generate synthetic data during training.
Generative-based methods perform well on datasets with sufficient training data so the feature generators (usually GAN~\cite{li2019leveraging,vyas2020leveraging} or VAE~\cite{verma2018generalized}) can be fully optimized. On the other hand, embedding-based methods try to learn an embedding function connecting visual features with corresponding semantic features without the use of feature generators.
Emebdding-based methods consist of many sub-categories, including graph-based~\cite{zhao2017zero}, meta learning-based~\cite{hu2018correction}, attention-based~\cite{xu2020attribute,huynh2020fine,liu2020attribute,liu2021goal,chen2022msdn}, \etc Our work can be categorized as the attention-based. 
Attention-based methods focus on discriminative regions of images and encode them into local features. 
The attention mechanism is usually implemented explicitly~\cite{xu2020attribute,huynh2020fine,liu2021goal,chen2022msdn} but can also be simulated from the middle layers of the neural network~\cite{Yang2021OnIA}.   

Our work learns both class and attribute representations for ZSL. Before our work, 
class representations have been exploited in several works \cite{jiang2018learning,zhang2017learning,Zhang2020APZ}, where they learn class prototypes from class semantics and optimize them against visual features from images. We follow these works to utilize class representations. Our difference to them mainly lies in the network design where our new framework enables a joint optimization of class and attribute representations in ZSL. As for the attribute representation, \cite{xu2020attribute,liu2021goal} make use of attribute features in images. However, their attribute features are extracted from each individual image and may not be representative enough as a consequence of the intra-attribute variance. We for the first time explicitly learn attribute prototypes beyond images and optimize them with attribute-level features within images.
Moreover, \cite{xu2020attribute,liu2021goal} learn a visual-to-semantic mapping while ours is the opposite. The former, according to~\cite{shigeto2015ridge}, is more likely to face the hubness problem.

\subsection{Contrastive learning} 
The goal of contrastive learning is to learn an embedding space in which similar samples are pushed close and dissimilar ones are pulled away~\cite{le2020contrastive}.  It can be applied to either labeled~\cite{khosla2020supervised} or unlabeled data~\cite{he2020momentum}. The latter is getting very popular in the self-supervised learning, where the essential idea is to enforce the embedding of the same sample of multiple views to be similar. 
Contrastive learning is recently used in ZSL by~\cite{han2021contrastive}. It introduces class-level and instance-level contrastive losses into a generative-based model. The contrastive loss in our work is applied to attribute-level features in an embedding-based model. Unlike~\cite{han2021contrastive} using all the positive/negative samples for each anchor to construct the loss, we introduce a new hard example-based contrastive loss which uses only hard attribute-level features based on their attention peaks and mutual distances. {In addition, our loss forms are also different.}

\subsection{ Transformers}
Compared to the CNN-based attention architecture, the self-attention architecture in transformers has demonstrated to be superior in many natural language processing (NLP) tasks~\cite{vaswani2017attention,devlin2018bert,brown2020language}. Because of its huge success in NLP, many computer vision researchers also start using it. One  successful example is the vision transformer~\cite{dosovitskiy2020image} (ViT): its essential idea is to slice the image into a sequence of patches and treat their embeddings as patch tokens; an extra classification token is also added to the transformer to generate global features for image classification. 

A recent work, ViT-ZSL~\cite{alamri2021multi}, has tried to adapt the ViT into ZSL as a backbone for feature extraction. ~\cite{alamri2021multi} replaces the classification head in ViT with a FC layer to project the global feature into the semantic space for ZSL. We instead use the global feature directly in the visual space. Also, the patch embedding in ViT is dropped in~\cite{alamri2021multi}, while we use it in an attention-based attribute localization scheme for attribute-level feature embedding. Overall, we introduce a different way to adapt ViT as a backbone in our work.

\section{Method}
\begin{figure*}[t]
\centering
\includegraphics[width=1\textwidth]{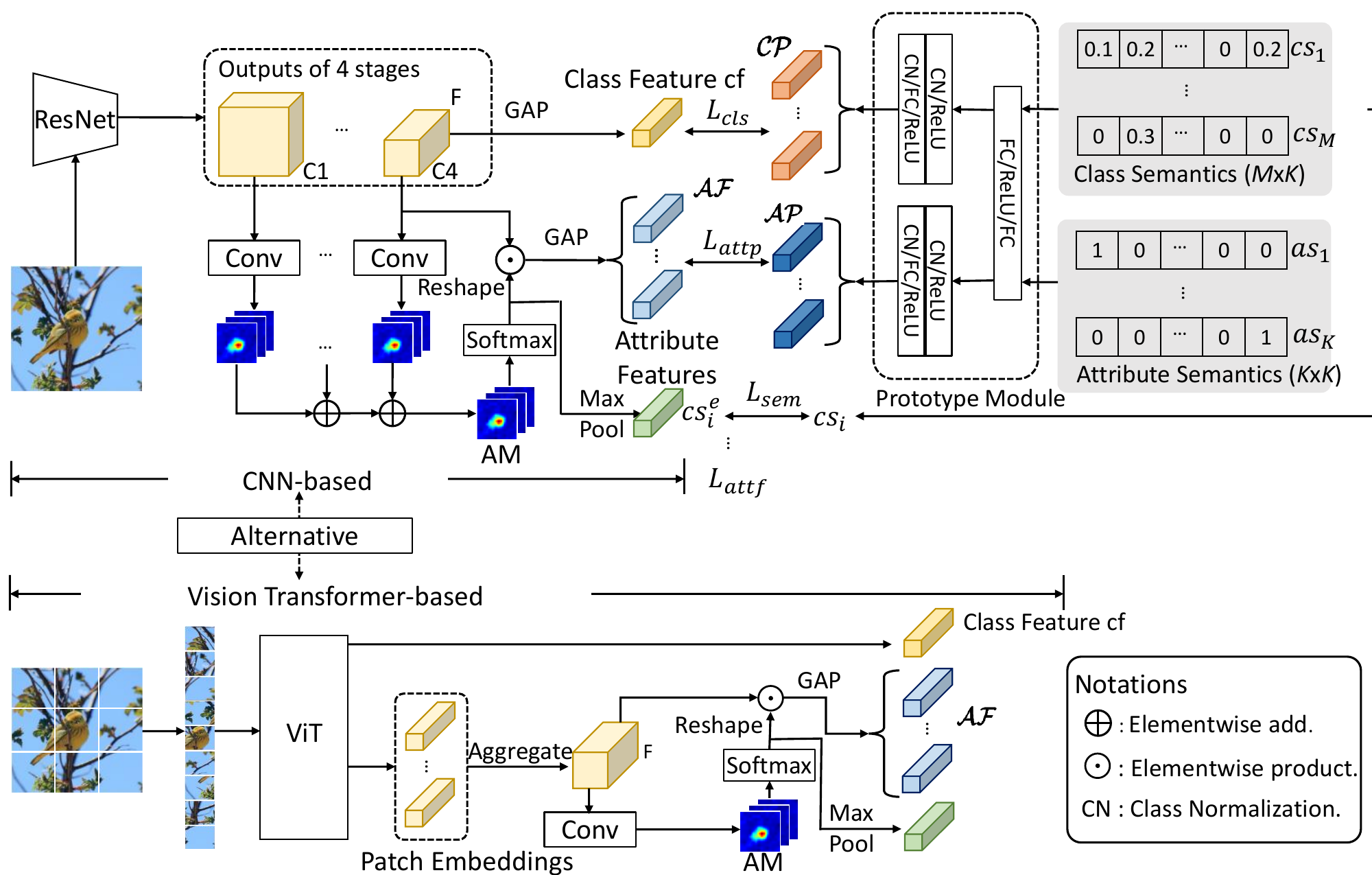}
  \caption{Framework of our CoAR-ZSL. Looking from the right, the prototype module takes the input of class and attribute semantics and outputs the class and attribute prototypes. Looking from the left, for an input image, class- and attribute-level features will be extracted separately to contrastively optimize with corresponding class and attribute prototypes ($\mathcal L_\text{attp}$). Another hard example-based {contrastive loss} ($\mathcal L_\text{attf}$) is also included to reinforce the attribute representation learning; it is not specified in the figure, see Sec.~\ref{sec:attrep} for more details. 
  }
\label{fig:zeroshot}
\end{figure*}

\subsection{Problem setting}\label{sec:setting}
Zero-shot learning aims to recognize classes that have no labeled data in the training set. 
We follow other embedding-based methods~\cite{xie2019attentive,xie2020region,chen2022msdn} to use the seen class semantics and their visual features to learn an embedding function during training.   
 
We denote by ${\mathcal X}^s$ and ${\mathcal X}^u$ the set of image visual features, $\mathcal Y^s$ and $ \mathcal Y^u$ the set of image class labels, 
for seen ($s$) and unseen ($u$) classes, respectively. 
The training set for seen classes is represented as $\mathcal D^s=\{x^s_i,y^s_i\}_{i=1}^{N^s}$ of $N^s$ elements and has $M^s$ classes in total.  
The test set for unseen classes is similarly represented as $\mathcal D^u=\{x^u_i,y^u_i\}_{i=1}^{N^u}$ for $N^u$ elements and has $M^u$ classes in total. There are in total $K$ object attributes shared between seen and unseen classes while there is no overlap between $\mathcal D^s$ and $\mathcal D^u$. 
In the conventional ZSL, the task is to recognize only unseen classes. 
In the more realistic GZSL, the task is to recognize both the seen and unseen classes. 
Unless specified, below we omit superscript $s$ or $u$.

\subsection{Overview}\label{sec:overview}
The overview of our proposed new framework CoAR-ZSL is shown in Fig.~\ref{fig:zeroshot}: two main data streams flow into it from the right and left side for prototype generation and feature embedding, respectively. For feature embedding, two alternative backbones, \ie CNN-based (default) and Transformer-based, are presented.

\subsubsection{Prototype generation} Looking from the right, we design a class and attribute prototype generation module which takes the input of class and attribute semantics, $\mathcal {CS} = \{cs_i\}_{i=1}^M$ and $\mathcal {AS} = \{as_j\}_{j=1}^K$, and outputs class and attribute prototypes, $\mathcal {CP} = \{cp_i\}_{i=1}^M$ and $\mathcal {AP} = \{ap_j\}_{j=1}^K$, respectively. 

\subsubsection{Class representation learning} 
Given an input image $x$ from the left side, we can extract its class-level global feature $cf$ directly from the backbone. Its cosine similarity is computed against $\mathcal {CP}$ and optimized via the cross entropy loss for classification ($\mathcal L_\text{cls}$).

\subsubsection{Attribute representation learning} Given the image $x$, we can extract its attribute-level local features $\mathcal {AF} = \{af_i\}_{i=1}^K$ via the attention-based attribute localization scheme, which produces an attention tensor $AM$ incorporating a set of attention maps, which are used as soft masks to localize
different attribute-related regions in $x$ and extract $\mathcal{AF}$ from them. For each $af$ in $\mathcal {AF}$, it is optimized against the corresponding $ap$ using a contrastive triplet loss ($\mathcal L_\text{attp}$). 
For learning better $af$ across images, another hard example-based contrastive optimization loss ($\mathcal L_\text{attf}$) is also devised to reinforce the similarity of  attribute-level features corresponding to the same attribute. Finally, to focus $AM$ on the attribute-related regions, we max-pool it to obtain a semantic vector $cs^e$ and minimize its L2 distance to the ground truth $cs^g$ ($\mathcal L_\text{sem}$).

\subsubsection{Inference} At testing, the input class-semantics for seen class are replaced by the corresponding semantics of unseen classes (for ZSL) or of all classes (for GZSL) to embed new sets of class prototypes. We compute the cosine similarities from the class-level feature of a test image to embedded class prototypes to decide the class label of the image.

\subsection{Class representation learning}
\label{sec:prototype}

Given the class semantics in $\mathcal {CS}$, each $cs_i$ is in the form of a $K$-dimensional vector indicating the presence/absence of the $K$ attributes in this class. The digits in the vector can be binary or continuous (ours is continuous). They are stored as a $M \times K$ matrix and fed into the prototype generation module (specified later) to obtain class prototypes $\mathcal {CP}$. Each $cp$ is of $C$ dimensions. 
On the other hand, the class-level feature $cf$ is easily obtained by applying global average pooling (GAP) on the backbone feature tensor $F$ ({Fig~\ref{fig:zeroshot}}).

Given $\mathcal {CP}$ and $cf$ for the input image $x$,  we can compute the cosine similarity $cos(cf,cp_i)$
from $cf$ to any class prototype $cp_i$. Assuming $x$ belongs to the $i$-th class, the predicted probability for this class is given by
\begin{equation} \label{eq:scaling}
  p_i=\frac{\exp(\alpha\cdot cos(cf,cp_i))}{\sum_{j = 1}^M\exp(\alpha\cdot cos(cf,cp_j)))},
\end{equation}
where $\alpha$ is the scaling factor. The {cross entropy} loss is utilized to optimize $p_i$:
\begin{equation}
\mathcal{L}_\text{cls}=-\log p_i~~.
\end{equation}

\subsection{Attribute representation learning}\label{sec:attention}

\subsubsection{Prototype generation} \label{sec:prototype}
We design a prototype generation module to perform the semantic-to-visual mapping from class and attribute semantics to their corresponding visual features. It is a multilayer perceptron (MLP) with a few shared layers and two identical branches for class and attribute prototype generation, respectively. The shared layers include two FC layers with the first followed by Relu. Each separate branch contains one FC, two Relus before and after the FC. Inspired by the class normalization (CN) in~\cite{skorokhodov2020class}, which was introduced to preserve the variance of semantic-to-visual mapping, we follow its design to have two CN before the FC in each branch. CN can be seen as BatchNorm without the affine transform. 

For attribute prototype generation, the input attribute semantics $\mathcal {AS}$ are set as one-hot vectors such that $as_j$ only has its $j$-th value being nonzero in the vector. $\mathcal {AS}$ can be seen as a basis to construct $\mathcal {CS}$. Other orthogonal bases may also work to distinguish attributes whilst the one-hot form works the best empirically. $\mathcal {AS}$ are stored as a $K \times K$ matrix and fed into the prototype generation module to obtain $\mathcal {AP}$. Each $ap$ is of $C$ dimensions. 

Notice class prototype generation also utilizes this module which takes the input of $\mathcal {CS}$ and outputs $\mathcal{CP}$. 

\subsubsection{Attribute-level feature embedding.} 
 For the extraction of attribute-level features, we follow~\cite{xie2020region,xie2019attentive} to adopt an attention-based mechanism. By integrating it into our backbone architecture, we present the attention-based attribute localization scheme. 

We first add four convolution layers after all the four stages of the backbone (ResNet101, C1$\sim$C4) respectively to extract feature tensors of the same resolution $H \times W \times K$ over multiple scales. These attention tensors $AM^s$ are pixel-wisely added together to produce the $AM$, $AM = \sum_{s=1}^4 {AM}^s$. We apply the Softmax function on each feature map of $AM$ to move it in the range 0 to 1. The feature map $am_j$ in $AM$ hence serves as a soft mask indicating the potential localization of the $j$-th attribute. To obtain attribute-level features, we apply the bi-linear pooling between $AM$ and the image feature tensor $F$. It is basically to apply each attribute-level soft mask to feature maps in $F$ to localize this attribute~\cite{lin2015bilinear}. {Specifically, each $am_j$ is pixel-wisely multiplied to every feature map $f_i$ in $F$ followed by average pooling.} There are $C$ feature maps in $F$, which ends up with a $C$-dimensional feature vector $af_j$ for the $j$-th attribute. The process can be written as,      

\begin{equation}\label{eq:bilinear}
    af_j = \mathrm {GAP}(\mathcal R(Softmax(am_j)) \odot F),
\end{equation}
{where $Softmax(\cdot)$ is applied to the 2-dimensional ($H \times W$) input. We use $\mathcal R(\cdot)$ to expand $Softmax(am_j)$ (by replication) to $C$ channels so as to pixel-wisely multiply with $F$;} $\odot$ is the element-wise product. By iterating $am_j$ over $j$, we obtain the final $\mathcal {AF} =\{af_j\}_{j=1}^{K}$. \miaojing{To guarantee that $af_j$ is related to the $j$-th attribute, below we introduce the attribute representation optimization to optimize $\mathcal{AF}$ in the visual and semantic space respectively. }  
\subsubsection{Attribute representation optimization} \label{sec:attrep}
Attribute-level features $\mathcal {AF}$ are extracted within each individual image, the features for the same attribute can vary in forms across many images. Attribute prototypes, on the other hand, are generated beyond images from orthogonal attribute semantics. In order for the attribute representation optimization, we present two contrastive losses to let 1) any attribute-level feature be close to its corresponding attribute prototype; and 2) any attribute-level feature be close to other attribute-level features belonging to the same attribute. 

\indent \emph{Contrastive optimization of attribute-level features against attribute prototypes.} Given the set of attribute-level features $\mathcal {AF}$ for the input image $x$, we first filter out those attribute-level features $af_j$ whose corresponding attention maps $am_j$ have low peak values, \ie $\max_{a,b} (am_j(a,b)) < T$ ($T = 9$);  
these attention maps normally fail to localize corresponding attributes. 
\miaojing{
It is possible that this filtering step may remove $af_j$ whose corresponding $am_j$ has insignificant peak response, whilst the object in the image contains the $j$-th attribute.  This however will not cause a problem for network training: according to~\cite{guo2018zero}, attributes have different importance for the discrimination of a certain object class, only a subset of attributes are necessary for the visual recognition in ZSL.}

For a batch of images, we denote the eligible set of attribute-level features after filtering as ${\mathcal{\widetilde{AF}}} = \{\widetilde{af_j}\}_{j=1}^{\tilde{K}}$. Given any $\widetilde{af_j}$, we want to optimize it to be close to its corresponding attribute prototype $ap_j$ and be far from the other attribute prototypes $ap_{j^\prime\neq j}$ in the embedding space. A triplet loss function suits this purpose: 
\begin{equation}
\label{eq:triplet}
\mathcal{L}_\text{attp} = \sum_{j=1}^{\tilde{K}} |d(\widetilde{af_j}, ap_j) - \beta\min\limits_{j^\prime\neq j}d(\widetilde{af_j},ap_{j^\prime}) |_+
\end{equation} 
where $d(\cdot,\cdot)$ is the cosine distance (one minus cosine similarity), {$\beta$ is a hyperparameter to control the extent of pulling $d(\widetilde{af_j}, ap_j)$ away from $\min d(\widetilde{af_j},ap_{j^\prime})$. $|\cdot|_+$ denotes the Relu function which enforces $d(\widetilde{af_j}, ap_j)$ to be smaller than the minimal distance (times $\beta$) from $af_j$ to other $ap_j^\prime$.} 

\indent \emph{Hard example-based contrastive optimization of attribute features across images.} 
Besides $\mathcal{L}_\text{attp}$, we introduce a new hard-example based contrastive optimization loss to reinforce the attribute representation learning: it pulls attribute-level features across images corresponding to the same attribute to be close; corresponding to different attributes to be away.    

Similar to above, instead of using all attribute-level features, we only keep those with high-peak values, \ie ${\mathcal{\widetilde{AF}}}$. Further more, for every $\widetilde{af_j}$ in ${\mathcal{\widetilde{AF}}}$, we have other features corresponding to the same attribute to $\widetilde{af_j}$ as its positives; we keep only hard positives $\{\widetilde{af_{ju}^+}\}_{u=1}^U$ whose cosine similarities to $\widetilde{af_j}$ are smaller than $t$ ($t = 0.8$). Similarly, hard negatives $\{\widetilde{af_{jv}^-}\}_{v=1}^V$ are those features who correspond to different attributes to $\widetilde{af_j}$ and whose cosine similarities to $\widetilde{af_j}$ are larger than $1-t$. {The reason of using hard examples is  to leave certain space for the intra-attribute variation.} 
We use the {SupCon loss~\cite{khosla2020supervised}} for the objective function: 

\begin{equation}
\label{eq:tau}
\begin{split}
&    \mathcal{L}_\text{attf}=- \frac{1}{\tilde{K}}\sum_{j=1}^{\tilde{K}}  {\log \widetilde{p_{j}}}\\
    & \widetilde{p_{j}} =  \frac{\sum\limits_{u=1}^U\exp (\cos(\widetilde{af_{j}}, \widetilde{af_{ju}^+})/\tau)}{\sum\limits_{u=1}^U\exp(\cos(\widetilde{af_j},\widetilde{af_{ju}^+})/\tau)+\sum\limits_{v=1}^V\exp(\cos(\widetilde{af_j},\widetilde{af_{jv}^-})/\tau)}
\end{split}
\end{equation}
\noindent where we iterate $\widetilde{af_j}$ in ${\mathcal{\widetilde{AF}}}$ and average the loss values; $\tau$ is a scalar temperature.

\medskip

$\mathcal{L}_\text{attp}$ and $\mathcal{L}_\text{attf}$ are defined in the visual space as a result of the semantic-to-visual mapping (from $as$ to $ap$). Apart from them, we follow \cite{xu2020attribute,liu2021goal} to define another loss in the semantic space as a result of the visual-to-semantic mapping (from $am$ to $cs^e$). This loss helps different channel maps in $AM$ focus on different  attribute-related regions in image $x$: we apply Softmax and max-pooling to each feature map of $AM$ and get a vector $cs^e\in \mathbb{R}^{1 \times 1 \times K}$ in the semantic space whose $j$-th value indicates the maximum response of $j$-th attribute in the image. Assuming $x$ belongs to the class $i$, we minimize the L2 distance between $cs_i^e$ and the ground truth class semantics $cs_{i}^g$: 
\begin{equation}
\mathcal{L}_{sem} = \|cs_i^e-cs_{i}^g\|^2.
\end{equation}

\begin{table*}[thb]
  \centering
    \caption{Datasets statistics in terms of granularity, attributes, classes, and data split. }
  \begin{tabular}{c|c|c|ccc|ccc}
  \toprule
  \multirow{2}{*}{Dataset}&\multirow{2}{*}{Granularity}&\multirow{2}{*}{\#Attribute}&\multicolumn{3}{c|}{\#Class}&\multicolumn{3}{c}{\#Image}\\
  &&&Total&Train&Test&Total&Train&Test\\
  \midrule
  CUB&fine&312&200&150&50&11788&7057&4731\\
  SUN&fine&102&717&645&72&14340&10320&4020\\
  AWA2&coarse&85&50&40&10&37332&23527&13795\\

  \bottomrule
  \end{tabular}

  \label{tab:datasets}
  \end{table*}

\subsection{Loss function}
The overall loss function for our framework is,  
\begin{equation}\label{eq:loss}
    \mathcal{L} = \widetilde{\mathcal{L}_\text{cls}}+\lambda_\text{attp}{\mathcal{L}_\text{attp}}
    + \lambda_\text{attf}\mathcal{L}_\text{attf} +\lambda_\text{sem}\widetilde{\mathcal{L}_\text{sem}}
\end{equation}
$\mathcal{L}_\text{cls}$ and $\mathcal{L}_\text{sem}$ are defined for one image, we use~~$\widetilde{}$~~to denote the corresponding average loss in one batch so as to match with $\mathcal{L}_\text{attp}$ and $\mathcal{L}_\text{attf}$. {$\lambda_\text{attp}$, $\lambda_\text{attf}$, $\lambda_\text{sem}$ are corresponding loss coefficients.}

\subsection{Alternative: Transformer-based architecture.} 
Inspired by the success of vision transformer (ViT)~\cite{dosovitskiy2020image}, we provide an alternative backbone for our framework using the ViT. This change affects the left part of the framework for class- and attribute-level feature embedding (see Fig.~\ref{fig:zeroshot} bottom): 
given the input image $x$, it is sliced into $P$ evenly squared patches of size $Q \times Q$. 
They are embedded via the transformer encoder to obtain the feature tensor $F \in \mathbb {R}^{Q \times Q \times C}$. Positional embedding is added to patch embedding to keep the position information. Unlike the CNN-based architecture, the class-level feature $cf$ is directly embedded by adding 
an extra learnable classification token [CLS]. For attribute-level features, we adopt a similar attention-based attribute localization scheme to the CNN-based architecture: $F$ is passed through a convolutional layer to produce the attended feature tensor $AM \in \mathbb R^{Q \times Q \times K}$, whose $j$-th channel map $am_j$ serves as a soft mask for the $j$-th attribute localization. $AM$ is bi-linearly pooled with $F$ (see (\ref{eq:bilinear})) to obtain the attribute-level features $\mathcal {AF}=\{af_j\}_{j=1}^{K}$.  
\section{Experiments}

\subsection{Dataset and evaluation metrics}

We evaluate our method on three most widely used datasets CUB~\cite{welinder2010caltech}, SUN~\cite{patterson2014sun} 
and AwA2~\cite{xian2018zero}, and follow the proposed train and test split in \cite{xian2018zero}.
The statistics of the datasets is summarized in \cref{tab:datasets}: Caltech-UCSD Birds-200-2011 (CUB)~\cite{welinder2010caltech} is the most fine-grained dataset with 312 attributes. {It contains 11788 images with 150 seen classes and 50 unseen classes.}
SUN~\cite{patterson2014sun} is a large scene dataset which has 14340 images with 645 seen classes and 72 unseen classes. There are 102 attributes in total. Compared with the former two, AwA2~\cite{xian2018zero} is a relatively coarse dataset with only 85 attributes but it has the 37332 images which is the most of three datasets.  It contains 40 seen classes and 10 unseen classes.

We report results in both ZSL and GZSL settings (Sec.~\ref{sec:setting}). In the ZSL setting, we only evaluate the performance on unseen classes and use the top-1 accuracy ($T1$) as the evaluation metric. 
In the GZSL setting, we evaluate the performance on both seen and unseen classes and follow ~\cite{xian2018zero} to use generalized seen accuracy ($ {Acc_S}$), generalized unseen accuracy (${Acc_U}$) and their 
generalized Harmolic Mean(${Acc_H}$) as evaluate metrics. The former two are top-1 accuracy for seen and unseen classes respectively while the last one is obtained by, 
\begin{equation}
{Acc_H}=\frac{2\cdot {Acc_U}\cdot {Acc_S}}{{Acc_U}+{Acc_S}}.
\label{eq:harmonic mean}
\end{equation}
${Acc_H}$ measures the inherent bias towards seen classes~\cite{rahman2018unified}, which is a more important metric in GZSL. 

\subsection{Implementation details}
\label{sec:implementation}

As a default backbone, we choose the CNN-based architecture, ResNet101~\cite{he2016deep}, which is pre-trained on ImageNet1k (1.28 million images, 1000 classes). 
The input image resolution is 448 $\times$ 448 and global feature $cf$ is of 2048 dimensions. 
For transformer-based architecture, the large variant of the vision transformer (ViT)~\cite{dosovitskiy2020image} is used, which is pre-trained on ImageNet21k (14 million images, 21,843 classes). The input image resolution is 224 $\times$ 224 and global feature $cf$ is of 1024 dimensions; the patch size is $16 \times 16$, such that there are 196 patch tokens in total.
The hidden size of prototype generation module is set to 1024. 
$\tau$ in (\ref{eq:tau}) is set to 0.4 for CUB, 0.6 for SUN and AwA2. {$\beta$ in (\ref{eq:triplet}) is set to 0.5 and  $\alpha$ in (\ref{eq:scaling}) is 25 for all datasets}.  $\lambda_\text{attp}$, $\lambda_\text{attf}$, $\lambda_\text{sem}$ in (\ref{eq:loss}) is set as 0.1, 1 and 1, respectively. We choose the SGD optimizer and set the momentum as 0.9, initial learning rate 0.001, and weight decay 0.0001. The learning rate is decayed every 10 epochs, with the decay factor of 0.5. 
All models are trained with synchronized SGD over 4 GPUs for 20 epochs with a mini-batch of 32. {The mini-batch is organized as a 16-way 2-shot episode following~\cite{liu2021goal} meaning that we sample 2 images per class for 16 classes within this mini-batch}. 

\miaojing{We follow ~\cite{xian2017zero} to use the validation set for hyperparameter searching, which is a disjoint set of unseen classes left out from the training set of CUB, AwA2 and SUN respectively. For each hyper-parameter, we perform line search on a few candidate points to find one point that works well for all datasets.}

\subsection{Comparisons to state of the arts}

\begin{table*}[hbt]
\caption{Comparision with state of the art on CUB, SUN and AwA2. We report top-1 accuracy (${T1}$) for ZSL, $Acc_U$, $Acc_S$, $Acc_H$ for GZSL.  {Methods using CNN-based and transformer-based backbones are compared separately. For the former, we mark the best and second best results of ${T1}$ and $Acc_H$ in red and blue, respectively. For the latter, we mark the best results with underlines in red. *indicates using the transformer-based backbone.}}
  \centering

  \begin{tabular}{c|cccc|cccc|cccc}
  \toprule
  \multirow{3}{*}{Method}&\multicolumn{4}{c|}{CUB}&\multicolumn{4}{c|}{SUN}&\multicolumn{4}{c}{AwA2}\\
  &ZSL&\multicolumn{3}{c|}{GZSL}&ZSL&\multicolumn{3}{c|}{GZSL}&ZSL&\multicolumn{3}{c}{GZSL}\\
  &$T1$&$Acc_U$&$Acc_S$&$Acc_H$&$T1$&$Acc_U$&$Acc_S$&$Acc_H$&$T1$&$Acc_U$&$Acc_S$&$Acc_H$\\
  \midrule
  \multicolumn{13}{c}{Generative methods}\\
  \hline

  OCD-CVAE ~\cite{keshari2020generalized}&60.3&44.8&59.9&51.3&63.5&44.8&\textcolor{red}{42.9}&\textcolor{red}{43.8}&71.3&59.5&73.4&65.7\\
  LsrGAN~\cite{vyas2020leveraging}&60.3&48.1&59.1&53.0&62.5&44.8&37.7&40.9&-&-&-&-\\

  CE-GZSL~\cite{han2021contrastive}&77.5&63.9&66.8&65.3&63.3&48.8&38.6&43.1&70.4&63.1&78.6&70.0\\
  FREE~\cite{chen2021free}&-&55.7&59.9&57.7&-&47.4&37.2&41.7&-&60.4&75.4&67.1\\
  E-PGN~\cite{Yu2020EpisodeBasedPG}&72.4&52.0&61.1&56.2&-&-&-&-&73.4&52.6&\textcolor{red}{83.5}&64.6\\
  GCM-CF~\cite{Yue2021CounterfactualZA}&-&61.0&59.7&60.3&-&47.9&37.8&42.2&-&60.4&75.1&67.0\\

  TransferIF~\cite{Feng2021TransferIF}&-&52.1&53.3&52.7&-&32.3&24.6&27.9&-&\textcolor{red}{76.8}&66.9&71.5\\
  
  HSVA~\cite{chen2021hsva}&-&52.7&58.3&55.3&-&48.6&39.0&43.3&-&56.7&79.8&66.3\\

  ICCE~\cite{kong2022compactness}&-&67.3&65.5&66.4&-&-&-&-&-&65.3&\textcolor{blue}{82.3}&\textcolor{blue}{72.8}\\

  \hline
  \multicolumn{13}{c}{Non-generative methods}\\
  \hline

  CN-ZSL~\cite{skorokhodov2020class}&-&49.9&50.7&50.3&-&44.7&\textcolor{blue}{41.6}&43.1&-&60.2&77.1&67.6\\
    APZ~\cite{Zhang2020APZ}&53.2&58.2&37.8&45.9&61.5&38.9&39.7&39.3&-&-&-&-\\
  DVBE~\cite{Min2020DomainAwareVB}&-&53.2&60.2&56.5&-&45.0&37.2&40.7&-&63.6&70.8&67.0\\
 AREN~\cite{xie2019attentive}&71.8&63.2&69.0&66.0&60.6&40.3&32.3&35.9&67.9&54.7&79.1&64.7\\

 LFGAA~\cite{liu2019attribute}&67.6&36.2&80.9&50.0&61.5&18.5&40.4&25.3&68.1&27.0&93.4&41.9\\
  RGEN~\cite{xie2020region}&76.1&60.0&{73.5}&66.1&63.8&44&31.7&36.8&\textcolor{blue}{73.6}&{67.1}&76.5&{71.5}\\
  DAZLE~\cite{Huynh2020FineGrainedGZ}&66.0&56.7&59.6&58.1&59.4&52.3&24.3&33.2&67.9&60.3&75.7&67.1\\
  APN~\cite{xu2020attribute}&72.0&65.3&69.3&67.2&61.6&41.9&34.0&37.6&68.4&56.5&78.0&65.5\\
  GEM~\cite{liu2021goal}&\textcolor{blue}{77.8}&64.8&\textcolor{blue}{77.1}&\textcolor{blue}{70.4}&62.8&38.1&35.7&36.9&67.3&64.8&77.5&70.6\\
  MSDN~\cite{chen2022msdn}&76.1&\textcolor{blue}{68.7}&67.5&68.1&\textcolor{blue}{65.8}&\textcolor{red}{52.2}&34.2&41.3&70.1&62.0&74.5&67.7\\ 
  \textbf{CoAR-ZSL (Ours)}&\textcolor{red}{79.2}&\textcolor{red}{70.9}&\textcolor{red}{77.3}&\textcolor{red}{74.0}&\textcolor{red}{66.7}&\textcolor{blue}{50.6}&38.0&\textcolor{blue}{43.4}&\textcolor{red}{74.1}&\textcolor{blue}{68.1}&{79.1}&\textcolor{red}{73.2}\\
  \hline
  ViT-ZSL*~\cite{alamri2021multi}&-&{{67.3}}&{{75.2}}&{{71.0}}&-&{{44.5}}&\textcolor{red}{\underline{55.3}}&{{49.3}}&-&{{51.9}}&90.0&{{65.8}}\\
GEM-ZSL*~\cite{liu2021goal}&78.1&\textcolor{red}{\underline{73.7}}&{{71.5}}&{{72.6}}&75.3&{{48.9}}&54.7&{{51.6}}&68.2&{{64.0}}&\textcolor{red}{\underline{90.2}}&{{74.9}}\\
  \textbf{CoAR-ZSL* (Ours)}&\textcolor{red}{\underline{79.9}}&72.5&\textcolor{red}{\underline{76.3}}&\textcolor{red}{\underline{74.4}}&\textcolor{red}{\underline{79.4}}&\textcolor{red}{\underline{68.3}}&{{55.0}}&\textcolor{red}{\underline{61.0}}&\textcolor{red}{\underline{78.7}}&\textcolor{red}{\underline{76.5}}&{{88.2}}&\textcolor{red}{\underline{82.0}}\\
  \bottomrule
  \end{tabular}
  
  \label{tab:compare-with-sota}
\end{table*}

In \cref{tab:compare-with-sota} we compare our CoAR-ZSL to recent state of the arts~\cite{keshari2020generalized,vyas2020leveraging,han2021contrastive,chen2021free,Yu2020EpisodeBasedPG,Yue2021CounterfactualZA,Zhang2020APZ,Feng2021TransferIF,chen2021hsva,kong2022compactness,skorokhodov2020class,Min2020DomainAwareVB,xie2019attentive,xie2020region,Huynh2020FineGrainedGZ,xu2020attribute,liu2021goal,chen2022msdn,alamri2021multi,liu2019attribute} in both ZSL and GZSL settings. 
We partition them into generative methods~\cite{keshari2020generalized,vyas2020leveraging,han2021contrastive,chen2021free,Yu2020EpisodeBasedPG,Yue2021CounterfactualZA,Feng2021TransferIF,chen2021hsva,kong2022compactness} and non-generative methods~\cite{skorokhodov2020class,Min2020DomainAwareVB,xie2019attentive,xie2020region,Huynh2020FineGrainedGZ,Zhang2020APZ,xu2020attribute,liu2021goal,chen2022msdn,alamri2021multi,liu2019attribute} following~\cite{xu2020attribute,liu2021goal}. Our CoAR-ZSL is a non-generative method.

\subsubsection{CNN-based architecture} All methods except for~\cite{alamri2021multi} use the same CNN-based ResNet101 backbone with our CoAR-ZSL. CoAR-ZSL significantly outperforms the state of the art CNN-based methods on most of the indicators. In particular with $T1$ and $Acc_H$, which are two important indicators for ZSL and GZSL, it achieves {79.2, 66.7, 74.1} for $T1$ and {74.0, 43.4, 73.2} for $Acc_H$ on CUB, SUN and AwA2, respectively. Also, we would like to point out that our method produces very good results on all three datasets while the previous best results are spread over different methods on the three datasets.

Unlike those generative-based methods~\cite{skorokhodov2020class,Min2020DomainAwareVB,xie2019attentive,xie2020region,Huynh2020FineGrainedGZ,Zhang2020APZ,xu2020attribute,liu2021goal,chen2022msdn,alamri2021multi,liu2019attribute}, our CoAR-ZSL does not need to synthesize virtual features of unseen classes nor do we need the semantics or visual features of unseen classes during training. If we do a fairer comparison to other non-generative methods~\cite{skorokhodov2020class,Min2020DomainAwareVB,xie2019attentive,xie2020region,Huynh2020FineGrainedGZ,Zhang2020APZ,xu2020attribute,liu2021goal,chen2022msdn,alamri2021multi,liu2019attribute}, our improvements over the state of the art are even more!!  
Overall, CoAR-ZSL is apparently the most competitive method. 

\miaojing{Besides the comparison on the accuracy, we also provide the analysis of computational efficiency between our method and the state of the art. Following~\cite{alwani2022decore}, we use FLOPs as the measurement of the computation cost, which is the number of floating-point operations by forwarding a single sample in the network. We compare our method to other methods, \ie LFGAA~\cite{liu2019attribute}, AREN~\cite{xie2019attentive}, GEM~\cite{liu2021goal} and APN~\cite{liu2020attribute}, as they have released their codes and we are all embedding-based methods. For all methods, the batch size is set to 1 and the backbones are finetuned. We report FLOPs on the CUB dataset in Table~\ref{tab:ablation-flops}: the FLOPs of our CoAR-ZSL is close to other methods. }

\begin{table}[t]
\caption{Comparison of FLOPs.}
\label{tab:ablation-flops}
  \centering
  \begin{tabular}{c|c}
  \toprule
  Method&FLOPs\\
  \midrule
LFGAA~\cite{liu2019attribute}&$3.20\times10^{10}$\\
AREN~\cite{xie2019attentive}&$3.15\times10^{10}$\\
APN~\cite{xu2020attribute}&$3.18\times10^{10}$\\
GEM~\cite{liu2021goal}&$3.13\times10^{10}$\\
CoAR-ZSL&$3.16\times10^{10}$\\
  \bottomrule
\end{tabular}

\end{table}

\subsubsection{Transformer-based architecture} When replacing the CNN-based architecture with the transformer-based architecture, the improvement of CoAR-ZSL* over CoAR-ZSL is impressive! {It raises $T1$ to 
79.9, 79.4, 78.7 and $Acc_H$ to 74.4, 61.0, 82.0 for the three datasets, respectively.} The improvements on SUN and AwA2 are particularly significant: {$12.7\%$ and $4.6\%$ on $T1$, $17.6\%$ and $8.8\%$ on $Acc_H$.} These improvements make CoAR-ZSL* significantly outperforming the recent work, ViT-ZSL~\cite{alamri2021multi}, who also utilizes ViT~\cite{dosovitskiy2020image}.
The ViT backbone is pretrained on ImageNet21k which is not comparable to the CNN-based backbone. To make a fair comparison, we replace the backbone of the state of the art method GEM-ZSL~\cite{liu2021goal} with the ViT backbone and report the results in \cref{tab:compare-with-sota}: GEM-ZSL*. Our CoAR-ZSL* significantly outperforms GEM-ZSL*. 

\medskip 
\miaojing{Overall, the better accuracy obtained by using the transformer-based architecture also comes with three shortcomings compared to that using the CNN-based architecture: 1) the occupied memory is bigger; 2) the inference time is longer; 3) the pretraining cost is higher. In Table~\ref{tab:ablation-computing} we compare the occupied GPU memory during training, the inference time, and the pretrained dataset size between the CNN- and transformer-based models; specifically, they are ResNet101 and ViT-L~\cite{dosovitskiy2020image} in our experiment. The statistics are collected by running both models on the CUB dataset. One can clearly see that the transformer-based model is more expensive to be pre-trained, learned, and deployed. }

\begin{table*}[bht]
\caption{Comparison of ResNet101 and ViT-L in terms of occupied GPU memory, inference time and pretrained dataset size.}
\label{tab:ablation-computing}
  \small
  \centering
  \begin{tabular}{c|c|c|c}
  \toprule
  Method&Occupied GPU memory (GB)&Inference time (s)&Pretrained dataset size (million)  \\
  \midrule
ResNet101&11.8&0.046&1.28\\
ViT-L&13.6&0.233&14.2\\
  \bottomrule
\end{tabular}

\end{table*}
 \begin{table}[hbt]
\caption{Ablation study of loss terms in (\ref{eq:loss}). 
}
\label{tab:ablation-loss}
  \small
  \centering
  \scalebox{0.9}{
  \begin{tabular}{c|cc|cc|cc}
  \toprule
  \multirow{2}{*}{Method}&\multicolumn{2}{c|}{CUB}&\multicolumn{2}{c|}{SUN}&\multicolumn{2}{c}{AwA2}\\
  &$T1$&$Acc_H$&$T1$&$Acc_H$&$T1$&$Acc_H$\\
  \midrule
  CoAR-ZSL &\textbf{79.2}&\textbf{74.0}&\textbf{66.7}&\textbf{43.4}&\textbf{74.1}&\textbf{73.2}\\
CoAR-ZSL w/o $\widetilde{\mathcal{L}_\text{sem}}$&77.0&71.0&66.4&41.8&72.4&72.9\\
  CoAR-ZSL w/o ${\mathcal{L}_\text{attp}}$&76.6&{71.1}&{65.6}&{41.4}&{73.1}&{71.1}\\
  CoAR-ZSL w/o $\mathcal{L}_\text{attf}$&77.4&70.0&65.0&41.4&72.9&72.2\\
  \midrule
  CoAR-ZSL w/  $\widetilde{\mathcal{L}_\text{cls}}$-only &75.8&69.8&64.0&40.7&69.7&70.0\\
 CoAR-ZSL w/ $\mathcal{L}_\text{attf}$-no-HS &78.1&71.8&65.5&42.5&73.7&72.7\\
  \bottomrule
\end{tabular}
}
\end{table}

\begin{table}[h]

\centering
\caption{Ablation study on prototype module (PM) structure.}
\label{tab:ablation-prototype-module}
         \scalebox{0.9}{%
 \begin{tabular}{c|cc|cc|cc}
  \toprule
  \multirow{2}{*}{Method}&\multicolumn{2}{c|}{CUB}&\multicolumn{2}{c|}{SUN}&\multicolumn{2}{c}{AwA2}\\
  &$T1$&$Acc_H$&$T1$&$Acc_H$&$T1$&$Acc_H$\\
  \midrule
  PM-v1 &76.2&71.3&65.7&41.1&70.5&71.3\\
  PM-v2 &74.1&69.1&64.5&40.9&67.9&70.3\\
  PM w/o CN &72.2&66.2&63.1&37.6&65.2&68.4\\
  PM &\textbf{79.2}&\textbf{74.0}&\textbf{66.7}&\textbf{43.4}&\textbf{74.1}&\textbf{73.2}\\
  \bottomrule
\end{tabular}
}
\end{table}
\begin{table}[h]

\centering
\caption{Ablation study on attribute semantics.}
\label{tab:ablation-semantics}
\begin{tabular}{c|cc|cc|cc}
  \toprule
  \multirow{2}{*}{Method}&\multicolumn{2}{c|}{CUB}&\multicolumn{2}{c|}{SUN}&\multicolumn{2}{c}{AwA2}\\
  &$T1$&$Acc_H$&$T1$&$Acc_H$&$T1$&$Acc_H$\\
  \midrule
  Rnd &74.6&69.7&63.9&40.3&67.8&69.8\\
  Rnd-ort &76.4&72.4&64.2&41.3&68.8&70.8\\
  One-hot&\textbf{79.2}&\textbf{74.0}&\textbf{66.7}&\textbf{43.4}&\textbf{74.1}&\textbf{73.2}\\
  \bottomrule
\end{tabular}
\end{table}

\subsection{Ablation study}
Ablation study {is on all three datasets} and we report $T1$ and $Acc_H$ for ZSL/GZSL, using the CNN backbone.

\subsubsection{Impact of attribute representation optimization}

\miaojing{We show the impact of attribute representation learning by removing $\widetilde{\mathcal{L}_\text{sem}}$, ${\mathcal{L}_\text{attp}}$, ${\mathcal{L}_\text{attf}}$ or all three losses.}

The results are in~\cref{tab:ablation-loss} where it clearly shows that $\widetilde{\mathcal{L}_\text{sem}}$, ${\mathcal{L}_\text{attp}}$, and  ${\mathcal{L}_\text{attf}}$ all help the performance and they are complementary. \miaojing{As a key contribution of our work, we for the first time distinguish the concepts between attribute prototypes and attribute-level features and introduce the contrastive optimization of attribute representations. The benefits of doing this are reflected in the two losses ${\mathcal{L}_\text{attp}}$ and $\mathcal{L}_\text{attf}$ where the former is to learn explicit and robust attribute prototypes while the latter is to reinforce attribute-level features. For instance, if we remove ${\mathcal{L}_\text{attp}}$,  it means the attribute prototypes will no longer be used in our model (the prototype generation module is now only responsible for generating class prototypes), we can see a clear performance drop on all three datasets in both ZSL and GZSL settings. }
\begin{figure*}[t]
     \centering
     \begin{minipage}{0.3\linewidth}
        \subfloat[]{ \includegraphics[width=1.\textwidth]{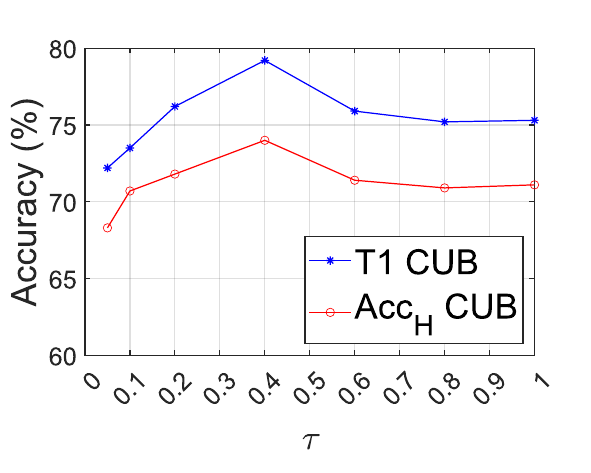}}
    \end{minipage}
    \begin{minipage}{0.3\linewidth}
         \subfloat[]{
         \includegraphics[width=1.\textwidth]{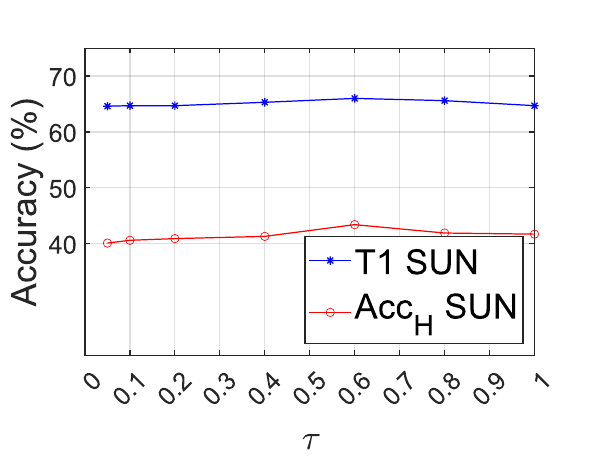}}
     \end{minipage}
     \begin{minipage}{0.3\linewidth}
          \subfloat[]{
            \includegraphics[width=1.\textwidth]{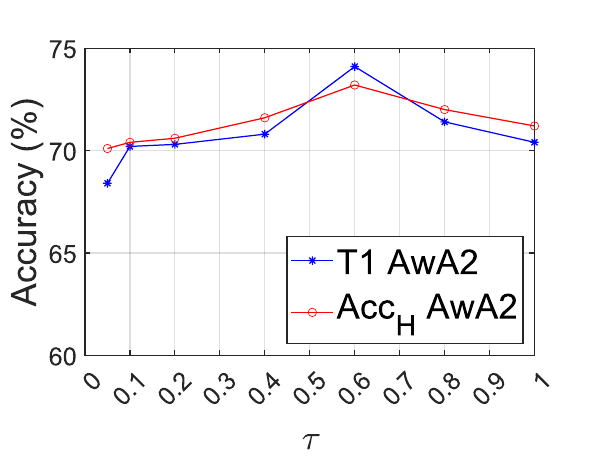}}
    \end{minipage}
        \caption{Parameter variation of hyperparameter $\tau$ in \cref{eq:tau}. }
        \label{fig:temperature}
\end{figure*}
\begin{figure*}[t]
     \centering
     \begin{minipage}{0.3\linewidth}
        \subfloat[]{ \includegraphics[width=1.\textwidth]{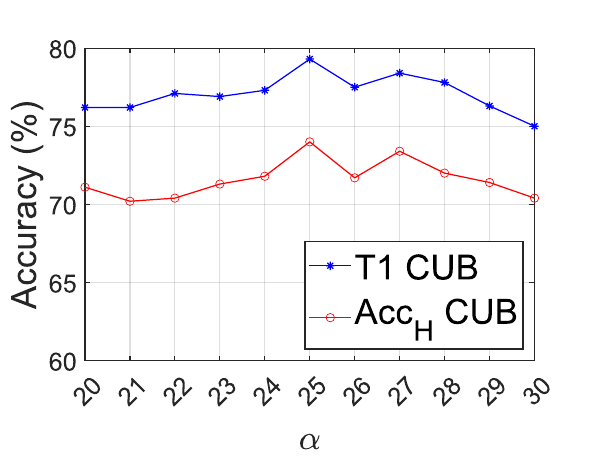}}
    \end{minipage}
    \begin{minipage}{0.3\linewidth}
         \subfloat[]{
         \includegraphics[width=1.\textwidth]{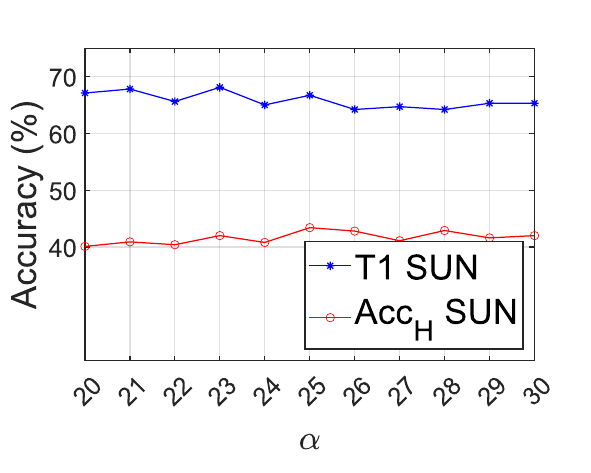}}
     \end{minipage}
     \begin{minipage}{0.3\linewidth}
          \subfloat[]{
            \includegraphics[width=1.\textwidth]{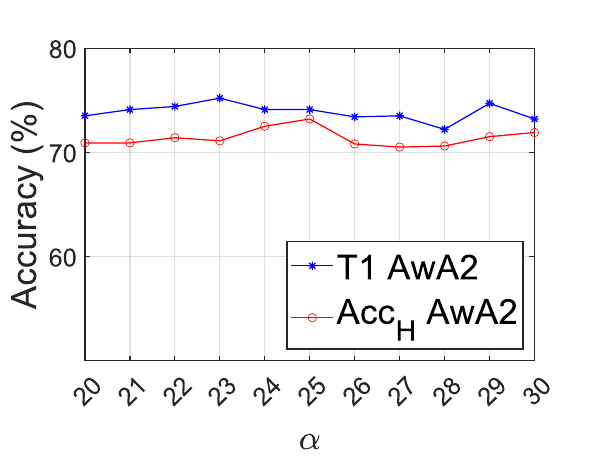}}
    \end{minipage}
        \caption{Parameter variation of hyperparameter $\alpha$ in \cref{eq:scaling}.}
        \label{fig:scale}
\end{figure*}

\begin{figure*}[h]
     \centering
     \begin{minipage}{0.3\linewidth}
        \subfloat[]{ \includegraphics[width=1.\textwidth]{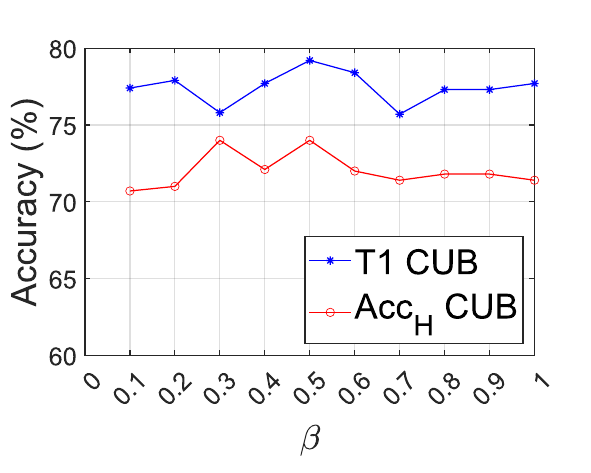}}
    \end{minipage}
    \begin{minipage}{0.3\linewidth}
         \subfloat[]{
         \includegraphics[width=1.\textwidth]{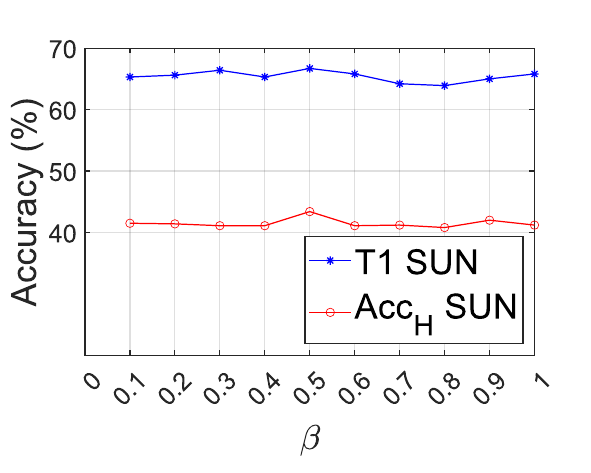}}
     \end{minipage}
     \begin{minipage}{0.3\linewidth}
          \subfloat[]{
            \includegraphics[width=1.\textwidth]{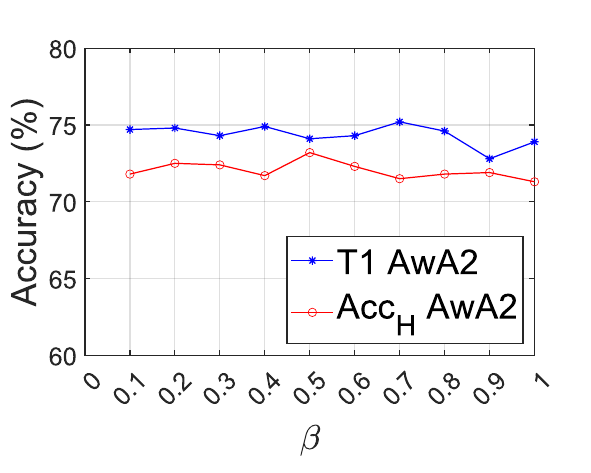}}
    \end{minipage}
        \caption{Parameter variation of hyperparameter $\beta$ in \cref{eq:triplet}.}
        \label{fig:beta}
\end{figure*}
\begin{figure*}[h]
     \centering
     \subfloat[\label{fig:CUB_T}]{
         \includegraphics[width=0.3\textwidth]{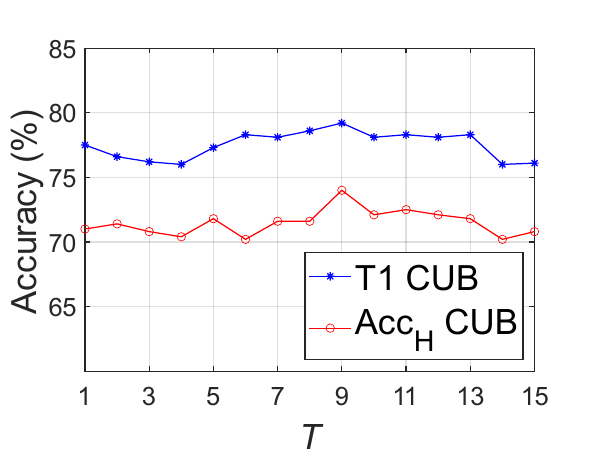}
       }
    \subfloat[\label{fig:SUN_T}]{
         \includegraphics[width=0.3\textwidth]{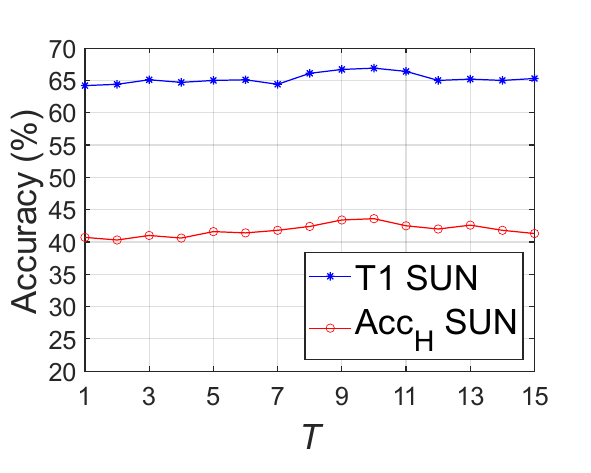}
        }
    \subfloat[\label{fig:AwA2_T}]{
         \includegraphics[width=0.3\textwidth]{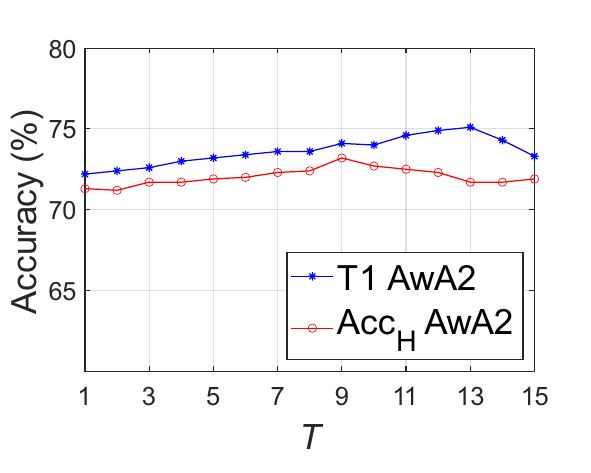}
      }

        \caption{Parameter variation of peak value threshold $T$.}
        \label{fig:peak value threhold}
\end{figure*}
\begin{figure*}[h]
     \centering
     \subfloat[\label{fig:lossweight1}]{
         \includegraphics[width=0.3\textwidth]{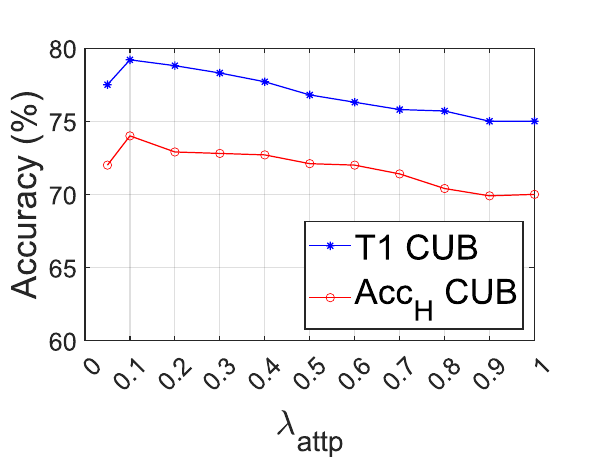}
       }
    \subfloat[\label{fig:lossweight2}]{
         \includegraphics[width=0.3\textwidth]{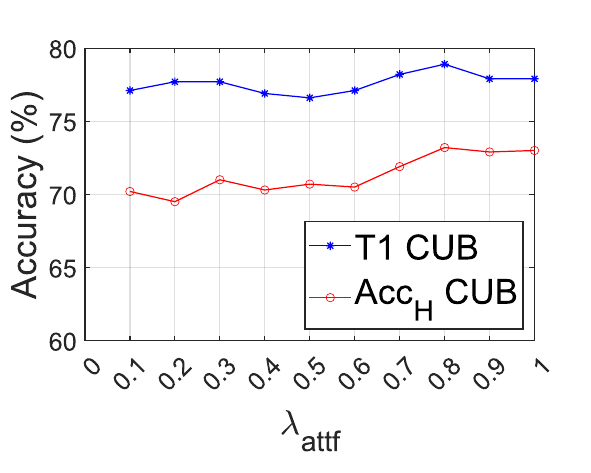}
      }
    \subfloat[\label{fig:lossweight3}]{
         \includegraphics[width=0.3\textwidth]{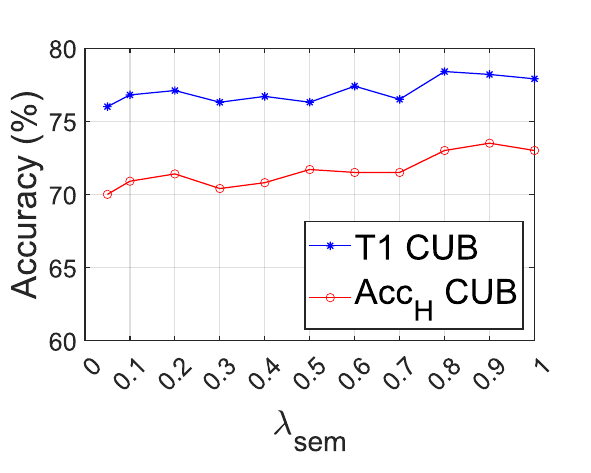}
        }
        \quad
         \subfloat[\label{fig:lossweight4}]{
         \includegraphics[width=0.3\textwidth]{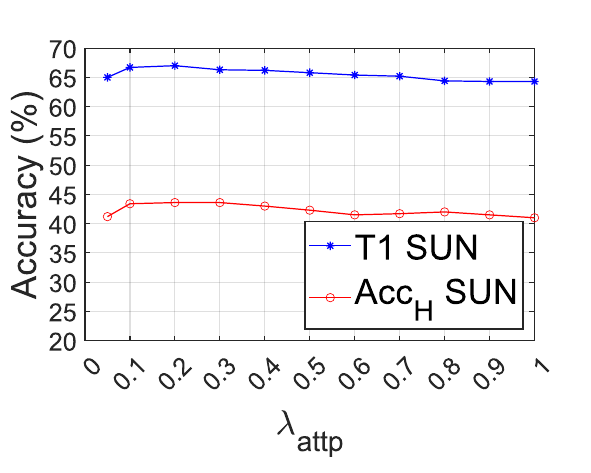}
      }
    \subfloat[\label{fig:lossweight5}]{
         \includegraphics[width=0.3\textwidth]{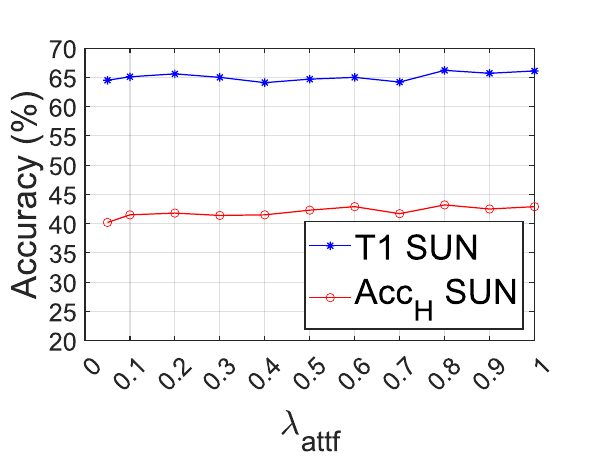}
      }
    \subfloat[\label{fig:lossweight6}]{
         \includegraphics[width=0.3\textwidth]{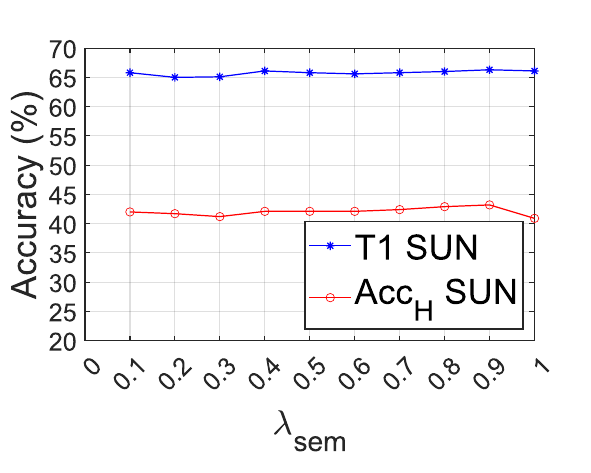}
        }
        \quad
         \subfloat[\label{fig:lossweight7}]{
         \includegraphics[width=0.3\textwidth]{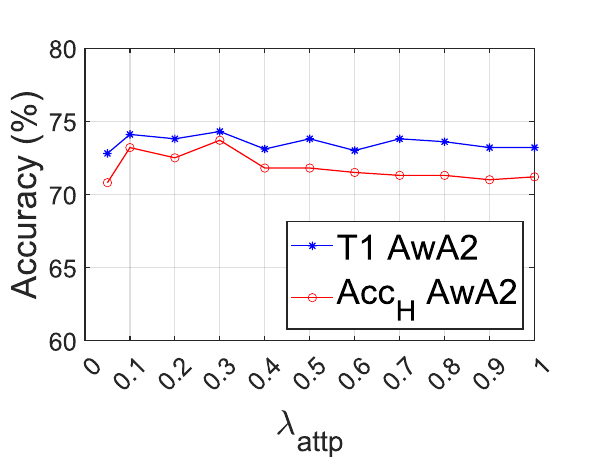}
      }
    \subfloat[\label{fig:lossweight8}]{
         \includegraphics[width=0.3\textwidth]{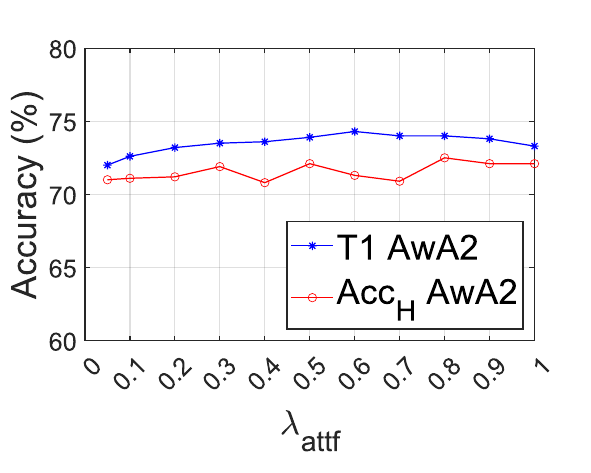}
      }
    \subfloat[\label{fig:lossweight9}]{
         \includegraphics[width=0.3\textwidth]{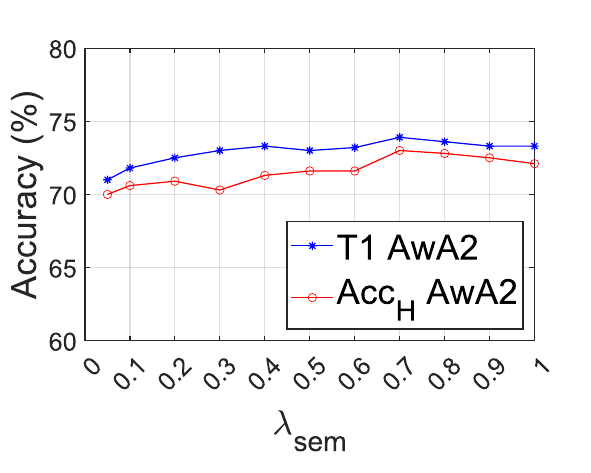}
        }
        \caption{Parameter variations for loss coefficients $\lambda_\text{sem}$, $\lambda_\text{attp}$ and $\lambda_\text{attf}$.}
        \label{fig:loss coefficient}
\end{figure*}

\emph{Hard examples in $\mathcal{L}_\text{attf}$.} We offer a variant of $\mathcal{L}_\text{attf}$ without hard example selection, denoting as $\mathcal{L}_\text{attf}$-no-HS. All examples will be used in (\ref{eq:tau}) and the result in Table~\ref{tab:ablation-loss} shows that CoAR-ZSL w/ $\mathcal{L}_\text{attf}$-no-HS performs clearly inferior to the original version of CoAR-ZSL.
This validates our idea of selecting hard examples in $\mathcal{L}_\text{attf}$. 

\subsubsection{Prototype generation module (PM)}
The module consists of a few shared layers and two identical branches for class and attribute prototype generation.  Two variants can be made on the module design: 1) we devise two identical modules without shared layers to generate prototypes for class and attribute separately (PM-v1); 2) we devise one module without separate branches to generate prototypes for class and attribute simultaneously (PM-v2). ~\cref{tab:ablation-prototype-module} shows that either case performs inferior to our original design. The shared part in our design enforces class- and attribute prototypes to share some common information while the separated part models their differences.
\begin{figure}[bht]
\centering
\subfloat[CNN-based architecture\label{fig:cnn-attribute-map}]{\includegraphics[width=1.\linewidth]{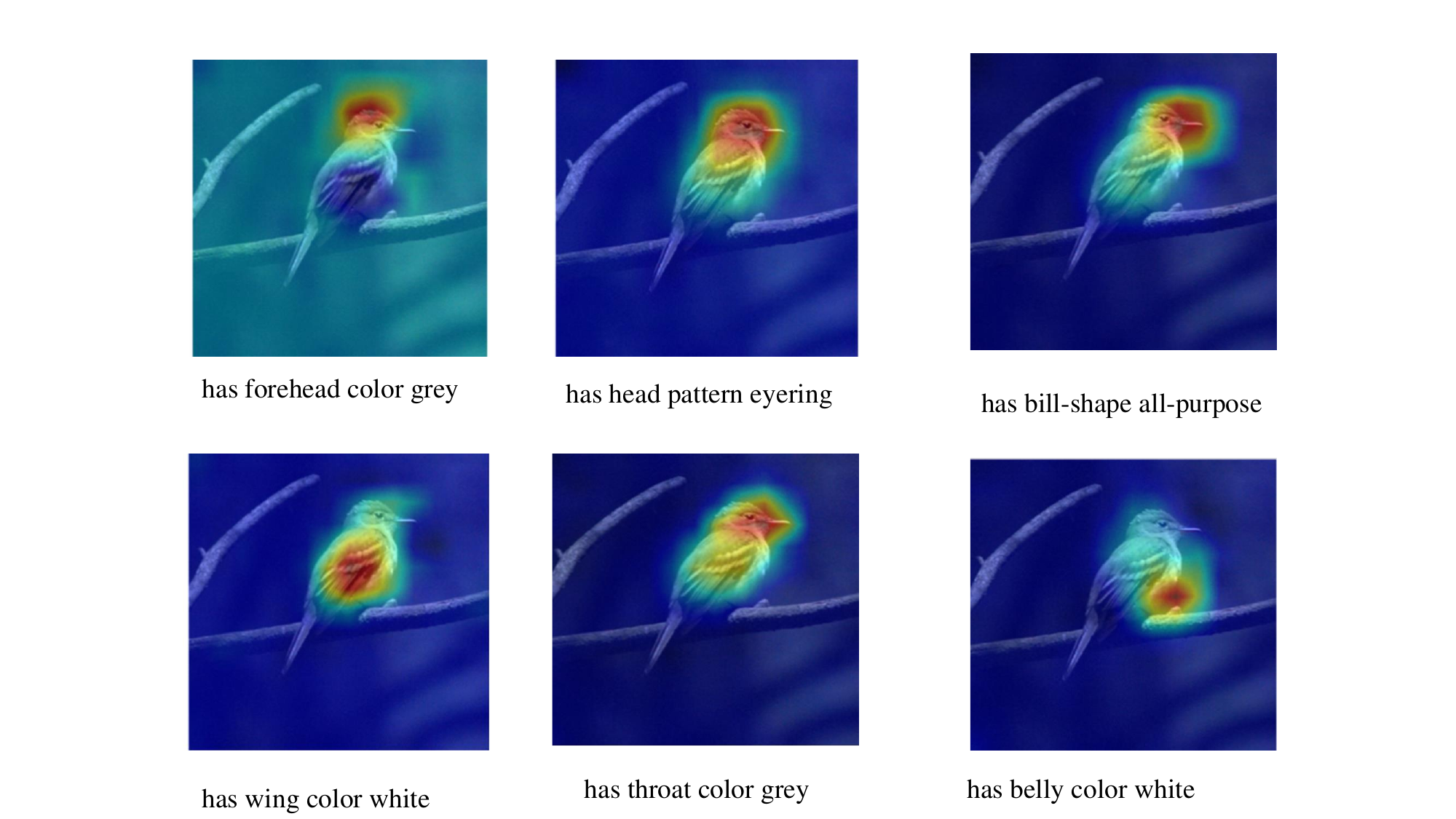}}
\hspace{15mm}
\subfloat[Transformer-based architecture\label{fig:vit-attribute-map}]{\includegraphics[width=0.9\linewidth]{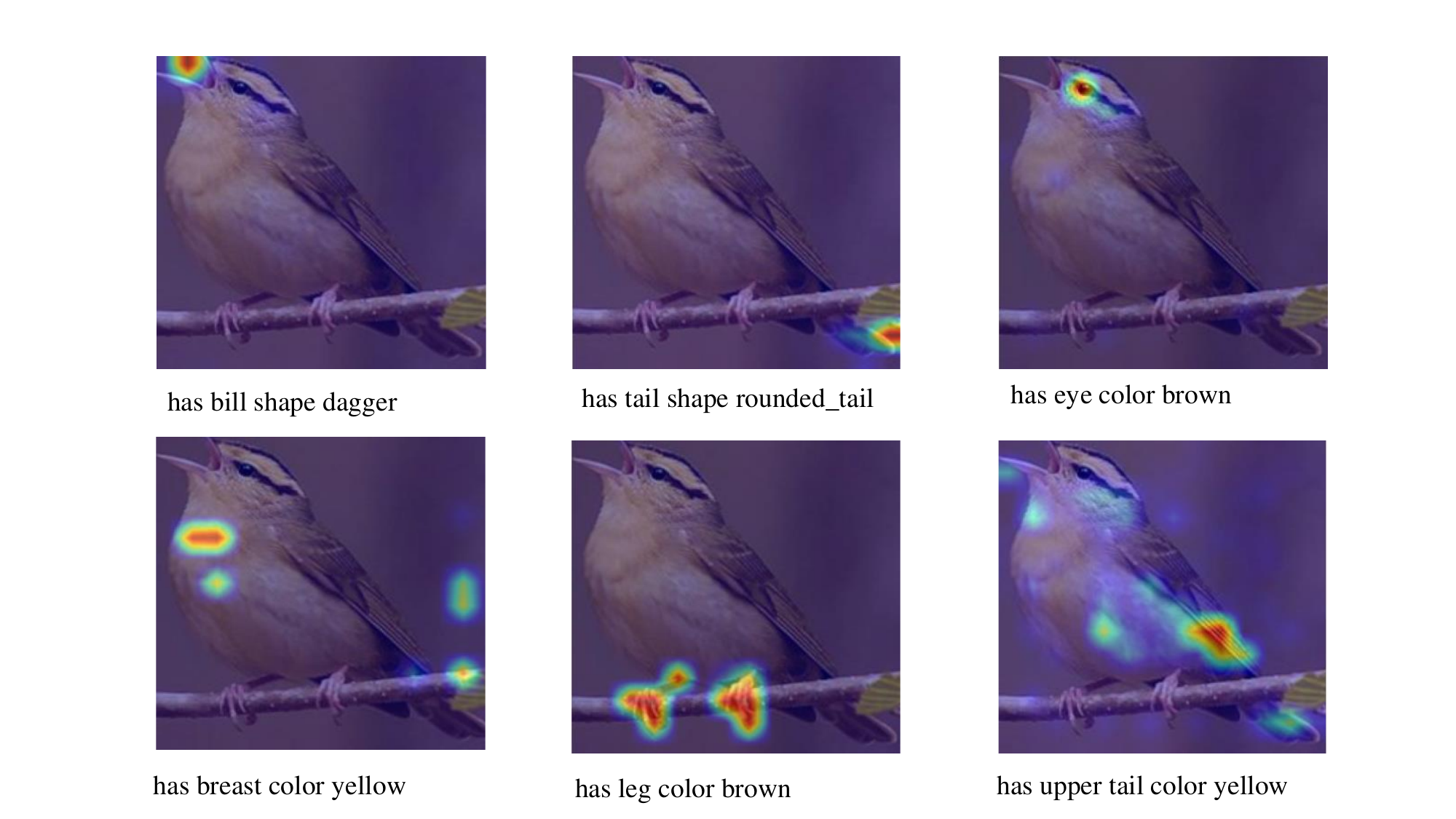}}
\caption{Localization of different attributes in an image.}
\label{fig:1img-multiatt}
\end{figure}

  \begin{figure*}[h] 
  \centering
  \includegraphics[width=0.85\linewidth]{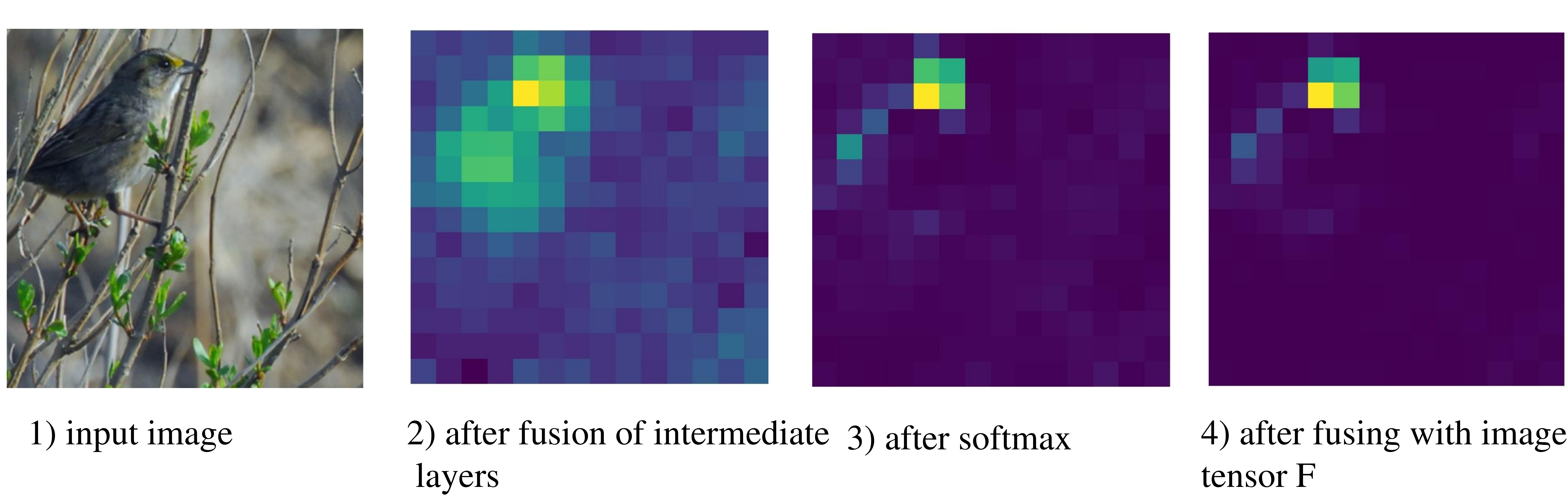}
  \caption{Visualization of intermediate feature maps in \cref{eq:bilinear}. }
  \label{fig:middlefeaturemap}
 \end{figure*}
The class normalization in the module (\cref{sec:prototype}) is also important. If we remove it (PM w/o CN), the performance drops substantially. \miaojing{Notwithstanding, we emphasize that our proposed attribute representation learning indeed plays a key role in the overall performance. As illustrated in Table~\ref{tab:ablation-loss}, with the loss {$\widetilde{\mathcal L_\text{cls}}$} only (CoAR-ZSL w/ $\widetilde{\mathcal L_\text{cls}}$-only), meaning that the attribute features and prototypes are no longer used in our model whilst CN is still applied in the prototype generation module for generating class prototypes, the performance is significantly dropped. CN is important for class representation learning, while the key contribution of this paper is attribute representation learning. }

\begin{figure*}
\centering
\subfloat[GEM\label{fig:tsne-gem}]{\includegraphics[width= 0.2\linewidth]{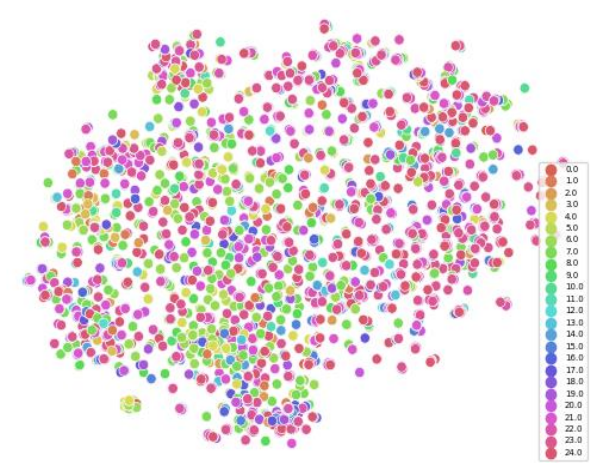}}
\hspace{2mm}
\subfloat[APN\label{fig:tsne-apn}]{\includegraphics[width=0.2\linewidth]{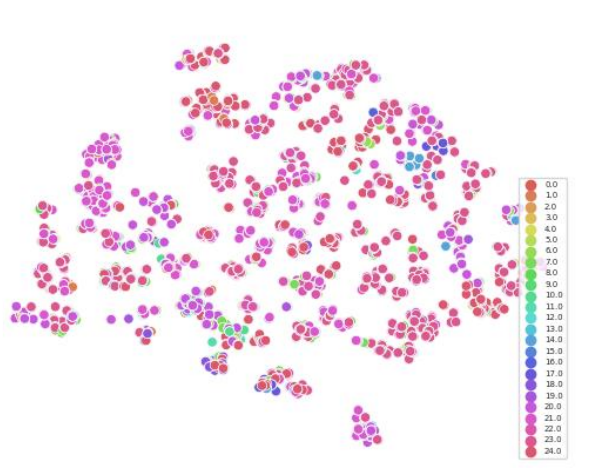}}
\hspace{2mm}
\subfloat[our CoAR-ZSL\label{fig:tsne-coar}]{\includegraphics[width=0.2\linewidth]{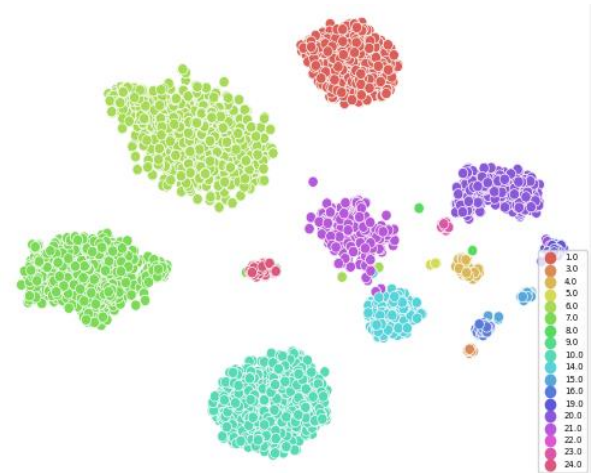}}
\hspace{2mm}
\subfloat[attribute feature distribution \label{fig:attr_dist}]{\includegraphics[width=0.2\linewidth]{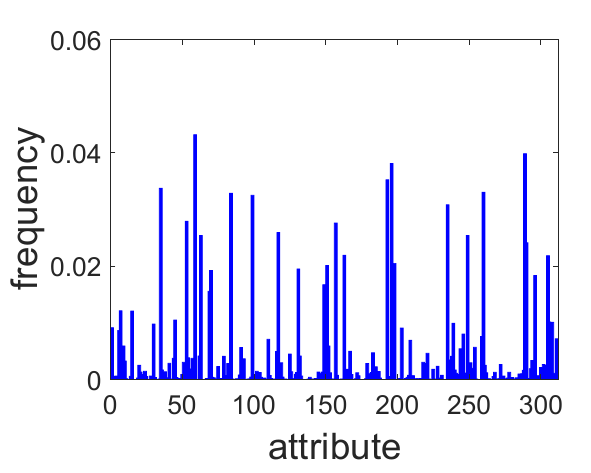}}
\caption{\miaojing{(a), (b) and (c) are the t-SNE visualization of attribute features for GEM~\cite{liu2021goal}, APN~\cite{liu2020attribute} and our CoAR-ZSL.} (d) is the attribute feature distribution for the CUB dataset. We plot the frequency of the features for each attribute.}
\label{fig:tsne}
\end{figure*}

\subsubsection{Attribute semantics}
We replace our one-hot attribute semantic vectors in $\mathcal{AS}$ with other forms to see the results. Specifically, a random orthogonal and non-orthogonal $K\times K$ matrices are made for the comparison.
For the random (non-orthogonal) matrix (Rnd), we generate it by drawing from a zero-mean Gaussian distribution with standard deviation of 0.1. For the orthogonal matrix (Rnd-ort), we use the Schmidt orthogonalization to orthogonalize the previous non-orthogonal matrix. The results are in~\cref{tab:ablation-semantics}. The one-hot form works the best and serves as the simplest and most suitable basis for the class semantics. Rnd-ort is better than Rnd as its orthogonality complies with the nature of distinct attributes.

\begin{table}[t]
\caption{Ablation study of extracting features from different positions of the CNN-based backbones.}
\label{tab:ablation-AM}
  \small
  \centering
  \begin{tabular}{c|cc|cc|cc}
  \toprule
  \multirow{2}{*}{Method}&\multicolumn{2}{c|}{CUB}&\multicolumn{2}{c|}{SUN}&\multicolumn{2}{c}{AwA2}\\
  &$T1$&$Acc_H$&$T1$&$Acc_H$&$T1$&$Acc_H$\\
  \midrule
  \multicolumn{7}{c}{ResNet101}\\
 \midrule
 4th stage&78.3&71.7&66.1&43.0&75.1&72.7\\
 2nd-4th stages&77.9&72.4&64.2&42.6&74.2&72.6\\
 3rd-4th stages&78.9&74.4&65.7&42.9&74.5&72.8\\
 1st-4th stages&\textbf{79.2}&\textbf{74.0}&\textbf{66.7}&\textbf{43.4}&\textbf{74.1}&\textbf{73.2}\\
  \midrule
  \multicolumn{7}{c}{VGG19}\\
  \midrule
  1st-4th blocks&65.1&63.3&59.9&39.2&68.5&67.6\\
  2nd-5th blocks&\textbf{68.1}&\textbf{63.4}&\textbf{60.0}&\textbf{40.3}&\textbf{69.5}&\textbf{69.7}\\
  \bottomrule
\end{tabular}

\end{table}

\begin{table}[t]

\centering
\caption{Comparison between using GAPs and GMPs.}
\label{tab:ablation-GAP}
\begin{tabular}{c|cc|cc|cc}
  \toprule
  \multirow{2}{*}{Method}&\multicolumn{2}{c|}{CUB}&\multicolumn{2}{c|}{SUN}&\multicolumn{2}{c}{AwA2}\\
  &$T1$&$Acc_H$&$T1$&$Acc_H$&$T1$&$Acc_H$\\
  \midrule
    GMP&75.5&70.5&57.5&32.7&68.3&68.6\\
  GAP&\textbf{79.2}&\textbf{74.0}&\textbf{66.7}&\textbf{43.4}&\textbf{74.1}&\textbf{73.2}\\
  \bottomrule
\end{tabular}

\end{table}

\subsubsection{Attention maps}
\miaojing{We use features from all four stages of ResNet101 (C1$\sim$C4) to obtain attention maps $AM$. It is a common practice to use features from these stages for object recognition task~\cite{lin2017feature}, as they contain different levels of details of the image, ranging from low-level visual information to high-level semantic information. To justify our design, we have also tried to extract features from different positions of ResNet101 and reported the results in Table~\ref{tab:ablation-AM}: it appears that using features from four stages works the best amongst all different combinations. These features are sufficient for accurately localizing attributes.}     

\miaojing{Besides ResNet, for other CNN-based backbones such as VGG19, we can follow~\cite{liu2019attribute} to use the last 4 blocks (VGG19 has overall 5 blocks) to extract $AM$. In Table~\ref{tab:ablation-AM}, we also present the results of using features from different blocks of VGG19. Apparently, using the 2nd-5th blocks is better than using 1st-4th blocks. } 

\subsubsection{GAP \vs GMP for feature tensor mapping}
\miaojing{We have utilized many global average pooling (GAP) operations to perform the mapping from feature tensors to feature vectors such that they can be later used for either classification or attribute representation learning. To achieve this goal, another popular choice is the global maximum pooling (GMP). GMP retains the most prominent local features of the feature map while GAP encourages the model to identify the complete extent of the object in the feature map. We adopt GAPs because in ZSL we need to pay attention to different details of objects corresponding to different attributes. In Table~\ref{tab:ablation-GAP}, we conduct an experiment for the comparison between using GAPs and GMPs on the CUB, SUN and AwA2 datasets in both ZSL and GZSL settings. Obviously, using GAP operations leads to better performance.}

\subsection{\bf Parameter variation}

\subsubsection{Temperature $\tau$ in \cref{eq:tau}}  
In \cref{fig:temperature}, we evaluate the effect of temperature $\tau$ in \cref{eq:tau}. We vary it from 0.05 to 1 on the three datasets and report the $T1$ and $Acc_H$ for ZSL and GZSL, respectively. It can be seen that the best performance occurs with $\tau$ equivalent to 0.4 and 0.6 for CUB and AwA2, respectively. The performance on SUN is rather stable by varying $\tau$. 
A low temperature will penalize more on hard negatives but a too low temperature will make no tolerance for outliers which can not be a good thing.
In practice, we set $\tau$ as 0.4 for CUB and $0.6$ for SUN and AwA2.

\subsubsection{Scaling factor $\alpha$ in \cref{eq:scaling}}
We show the impact of scaling factor $\alpha$ in \cref{eq:scaling} by varying it from 20 to 30 in \cref{fig:scale}. CoAR-ZSL achieves the best performance on CUB at $\alpha = 25$ for both $T1$ and $Acc_H$. For SUN and AwA2, the performance variation with different $\alpha$ is rather small on both $T1$ and $Acc_H$.
In practice, we set $\alpha$ as 25 for all datasets for simplicity.

\subsubsection{Ratio $\beta$ in \cref{eq:triplet}}
We also evaluate the effect of ratio $\beta$ in \cref{eq:triplet}. We plot the result in \cref{fig:beta}. CoAR-ZSL achieves the best performance on CUB at $\beta = 0.5$ for both $T1$ and $Acc_H$. For SUN and AwA2, the performance variation with different $\beta$ is rather small on both $T1$ and $Acc_H$. In practice, we set $\beta$ as  0.5 for all datasets for simplicity. 

\subsubsection{Peak value threshold $T$ in \cref{sec:attrep}}
\miaojing{We sweep $T$ from 1 to 15 for each dataset. It can be seen in Fig.~\ref{fig:peak value threhold} that 1) the best result for each dataset can indeed be slightly higher than that of using the default $T$ ($T = 9$) for all three datasets; 2) with the change of $T$, the result on the CUB varies while the results on the SUN and AwA2 are rather stable. This enables us to find one parameter that can work in general well for all three datasets, which is also good for the model's generalizability. }

\subsubsection{Loss weights $\lambda_\text{attp}$, $\lambda_\text{attf}$ and $\lambda_\text{sem}$}
\miaojing{For convenience, we perform line search on one loss weight while fixing the other two to 1. In Fig.~\ref{fig:loss coefficient}, we show the results by varying each loss weight from 0.05 to 1 on the CUB, AwA2 and SUN datasets. In general, a smaller $\lambda_\text{attp}$ compared to $\lambda_\text{attf}$ and $\lambda_\text{sem}$ (by one order of magnitude approximately) is better for the overall performance. Our default setting is that $\lambda_\text{attp}$ is 0.1 while the other two are 1. This, despite not being the optimum, is a fine setting for all three datasets. Theoretically, one could get better performance by optimizing loss weights via grid searching in each dataset, but this would be too costly and may not be beneficial for the model’s generalizability. }

\subsection{Qualitative results}
\subsubsection{{Attribute localization}} To visualize the {attention-based attribute localization} (Sec.~\ref{sec:attention}), we resize and normalize the attribute-related attention maps in $AM$ into the range [0,1] and draw them onto the original image in \cref{fig:1img-multiatt}. We show the examples using both the CNN-based and transformer-based architectures. In both figures, CoAR-ZSL can accurately locate attribute-related regions in images. 

\subsubsection{The intermediate feature maps in ~\cref{eq:bilinear}}
We illustrate the intermediate feature maps after fusion of intermediate layers, after softmax, and after fusion with the image feature tensor $F$. Referring to ~\cref{eq:bilinear}, they correspond to the feature map $am_j$, $Softmax(am_j)$, and $\mathcal R(Softmax(am_j)) \odot F$, respectively. 
We draw the attribute ``has bill shape dagger'' in \cref{fig:middlefeaturemap}: one can clearly observe how this attribute is localized in $F$ via the corresponding normalized soft mask ($Softmax(am_j)$). It aligns with the attribute on the bird's beak.    
  
\subsubsection{{t-SNE visualization of attribute features}} To validate the representativeness of the proposed attribute-level features, \miaojing{we draw the t-SNE of 4096 attribute-level features extracted over {multiple images} using methods of GEM~\cite{liu2021goal} (\cref{fig:tsne-gem}), APN~\cite{xu2020attribute} (\cref{fig:tsne-apn}) and our CoAR-ZSL (\cref{fig:tsne-coar}) . } For better visualization, we only plot the first 25 attributes in different colors.  
We can see that the attribute-level features in our model are clearly clustered according to the attribute they belong to while the ones in {GEM and APN} are rather mixed. Since we only keep attribute-level features with high peak values in their corresponding attribute-related attention maps ($AM$), it can be seen the kept features tend to form a number of big clusters for a number of attributes. This is indeed consistent to the attribute feature distribution shown in \cref{fig:attr_dist}: only a number of attributes have high frequencies.  The similar observation is made in \cite{pourpanah2020review} that many fine-grained classes contain discriminative information in a few regions.

\begin{figure}[t] 
  \centering{
  \includegraphics[width=0.9\linewidth]{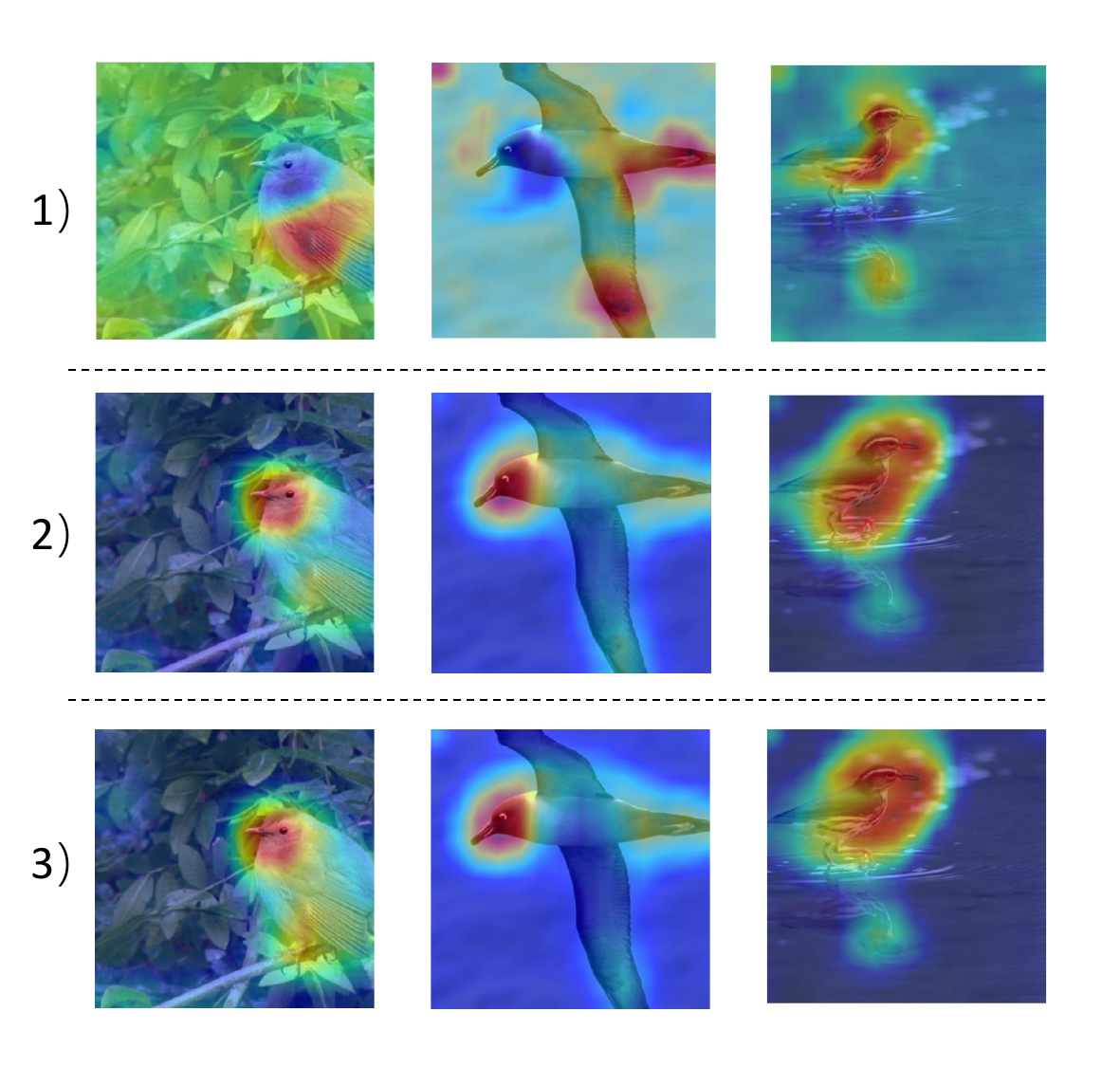}}
  \caption{Visualization of three types of activation maps: 1) adding up all attribute-related attention maps in $AM$ (first row); 2) adding up only high-response attribute-related attention maps in $AM$ (second row); 3) adding up all feature maps in $F$ (third row).
  }
  \label{fig:class syn}
\end{figure}

\subsubsection{Qualitative evidence for selecting high-response attention maps for constrastive optimization}
We plot three types of activation maps in \cref{fig:class syn} in three rows: 1) adding up all attribute-related attention maps in $AM$; 2) adding up only high-response attribute-related attention maps in $AM$; 3) adding up all feature maps in $F$. 
We know that each attention map $am_j$ in $AM$  signifies the feature response to one attribute $j$. 
Adding them together can be a reflection of these attributes on certain object class. On the other hand, since the class-level feature $cf$ is actually obtained from $F$, adding all feature maps in $F$ is indeed a reflection of the object class in the image. Having a look at \cref{fig:class syn}, activation maps in the second row are clearly cleaner and more object-focused than those in the first row, and are visually closer to those in the third row. This suggests that there exist noises in those low-response attention maps in $AM$ which could be brought into the activation map if adding them up. While those high-response attention maps in $AM$ represent the real attributes contained in the object class, adding these maps can end up with a proper reflection of attributes on this object class.

\section{Conclusion}
This paper proposes a novel embedding-based ZSL framework, CoAR-ZSL, for explicitly learning attribute representations. The representations include both image-specific features and image-agnostic prototypes for attributes. They are contastively optimized in the network to learn a robust classifier. A hard example-based contrastive learning scheme is also introduced to reinforce the learning of attribute representations. We utilize two backbones for CoAR-ZSL, CNN-based and Transformer-based, where the latter performs better than the former yet requires more computational cost. Extensive experiments on standard benchmarks demonstrate the superiority of our CoAR-ZSL over state of the art.

\miaojing{One shortcoming of our method is that it relies on human-crafted attributes given in the dataset and cannot be applied to datasets without explicit attribute information. This limits the generalizability of our method. Possible solution may be directly learning shared prototypes across classes to simulate the attributes.}

\miaojing{Besides above, there are also other places that can be improved in the future work: for instance, we now use hard samples for the contrastive learning of attribute features. This may be improved by using the graph neural network where attribute relations can be encoded into graph edges, so that attribute features can be optimized and refined via the message passing in the graph. 
}

\section*{Acknowledgements}
This project has received funding from the National Natural Science Foundation of China (61836004, 62236009) and the Fundamental Research Funds for the Central Universities.

 \bibliographystyle{IEEEtran}
 \bibliography{main.bib}

\begin{thebibliography}{10}
\providecommand{\url}[1]{#1}
\csname url@samestyle\endcsname
\providecommand{\newblock}{\relax}
\providecommand{\bibinfo}[2]{#2}
\providecommand{\BIBentrySTDinterwordspacing}{\spaceskip=0pt\relax}
\providecommand{\BIBentryALTinterwordstretchfactor}{4}
\providecommand{\BIBentryALTinterwordspacing}{\spaceskip=\fontdimen2\font plus
\BIBentryALTinterwordstretchfactor\fontdimen3\font minus
  \fontdimen4\font\relax}
\providecommand{\BIBforeignlanguage}[2]{{%
\expandafter\ifx\csname l@#1\endcsname\relax
\typeout{** WARNING: IEEEtran.bst: No hyphenation pattern has been}%
\typeout{** loaded for the language `#1'. Using the pattern for}%
\typeout{** the default language instead.}%
\else
\language=\csname l@#1\endcsname
\fi
#2}}
\providecommand{\BIBdecl}{\relax}
\BIBdecl

\bibitem{krizhevsky2012imagenet}
A.~Krizhevsky, I.~Sutskever, and G.~E. Hinton, ``Imagenet classification with
  deep convolutional neural networks,'' \emph{NeurIPS}, 2012.

\bibitem{simonyan2014very}
K.~Simonyan and A.~Zisserman, ``Very deep convolutional networks for
  large-scale image recognition,'' \emph{arXiv preprint arXiv:1409.1556}, 2014.

\bibitem{he2016deep}
K.~He, X.~Zhang, S.~Ren, and J.~Sun, ``Deep residual learning for image
  recognition,'' in \emph{CVPR}, 2016.

\bibitem{snell2017prototypical}
J.~Snell, K.~Swersky, and R.~Zemel, ``Prototypical networks for few-shot
  learning,'' in \emph{NeurIPS}, 2017.

\bibitem{yang2020restoring}
Y.~Yang, F.~Wei, M.~Shi, and G.~Li, ``Restoring negative information in
  few-shot object detection,'' in \emph{NeurIPS}, 2020.

\bibitem{sung2018learning}
F.~Sung, Y.~Yang, L.~Zhang, T.~Xiang, P.~H. Torr, and T.~M. Hospedales,
  ``Learning to compare: Relation network for few-shot learning,'' in
  \emph{CVPR}, 2018.

\bibitem{romera2015embarrassingly}
B.~Romera-Paredes and P.~Torr, ``An embarrassingly simple approach to zero-shot
  learning,'' in \emph{ICML}, 2015.

\bibitem{xian2017zero}
Y.~Xian, B.~Schiele, and Z.~Akata, ``Zero-shot learning-the good, the bad and
  the ugly,'' in \emph{CVPR}, 2017.

\bibitem{wang2019survey}
W.~Wang, V.~W. Zheng, H.~Yu, and C.~Miao, ``A survey of zero-shot learning:
  Settings, methods, and applications,'' \emph{ACM Transactions on Intelligent
  Systems and Technology}, vol.~10, no.~2, pp. 1--37, 2019.

\bibitem{he2020momentum}
K.~He, H.~Fan, Y.~Wu, S.~Xie, and R.~Girshick, ``Momentum contrast for
  unsupervised visual representation learning,'' in \emph{CVPR}, 2020.

\bibitem{chen2021exploring}
X.~Chen and K.~He, ``Exploring simple siamese representation learning,'' in
  \emph{CVPR}, 2021.

\bibitem{xu2020attribute}
W.~Xu, Y.~Xian, J.~Wang, B.~Schiele, and Z.~Akata, ``Attribute prototype
  network for zero-shot learning,'' in \emph{NeurIPS}, 2020.

\bibitem{liu2021goal}
Y.~Liu, L.~Zhou, X.~Bai, Y.~Huang, L.~Gu, J.~Zhou, and T.~Harada,
  ``Goal-oriented gaze estimation for zero-shot learning,'' in \emph{CVPR},
  2021.

\bibitem{yu2017transductive}
Y.~Yu, Z.~Ji, J.~Guo, and Y.~Pang, ``Transductive zero-shot learning with
  adaptive structural embedding,'' \emph{IEEE Transactions on Neural Networks
  and Learning Systems}, vol.~29, no.~9, pp. 4116--4127, 2017.

\bibitem{skorokhodov2020class}
I.~Skorokhodov and M.~Elhoseiny, ``Class normalization for (continual)?
  generalized zero-shot learning,'' in \emph{ICLR}, 2020.

\bibitem{zhang2017learning}
L.~Zhang, T.~Xiang, and S.~Gong, ``Learning a deep embedding model for
  zero-shot learning,'' in \emph{CVPR}, 2017.

\bibitem{chen2021hsva}
S.~Chen, G.-S. Xie, Y.~Liu, Q.~Peng, B.~Sun, H.~Li, X.~You, and L.~Shao,
  ``Hsva: Hierarchical semantic-visual adaptation for zero-shot learning,'' in
  \emph{NeurIPS}, 2021.

\bibitem{lampert2009learning}
C.~H. Lampert, H.~Nickisch, and S.~Harmeling, ``Learning to detect unseen
  object classes by between-class attribute transfer,'' in \emph{CVPR}, 2009.

\bibitem{church2017word2vec}
K.~W. Church, ``Word2vec,'' \emph{Natural Language Engineering}, vol.~23,
  no.~1, pp. 155--162, 2017.

\bibitem{vyas2020leveraging}
M.~R. Vyas, H.~Venkateswara, and S.~Panchanathan, ``Leveraging seen and unseen
  semantic relationships for generative zero-shot learning,'' in \emph{ECCV},
  2020.

\bibitem{xian2018feature}
Y.~Xian, T.~Lorenz, B.~Schiele, and Z.~Akata, ``Feature generating networks for
  zero-shot learning,'' in \emph{CVPR}, 2018.

\bibitem{chen2021semantics}
Z.~Chen, Y.~Luo, R.~Qiu, S.~Wang, Z.~Huang, J.~Li, and Z.~Zhang, ``Semantics
  disentangling for generalized zero-shot learning,'' in \emph{ICCV}, 2021.

\bibitem{xie2020region}
G.-S. Xie, L.~Liu, F.~Zhu, F.~Zhao, Z.~Zhang, Y.~Yao, J.~Qin, and L.~Shao,
  ``Region graph embedding network for zero-shot learning,'' in \emph{ECCV},
  2020.

\bibitem{liu2020attribute}
L.~Liu, T.~Zhou, G.~Long, J.~Jiang, and C.~Zhang, ``Attribute propagation
  network for graph zero-shot learning,'' in \emph{AAAI}, 2020.

\bibitem{chen2022msdn}
S.~Chen, Z.~Hong, G.-S. Xie, W.~Yang, Q.~Peng, K.~Wang, J.~Zhao, and X.~You,
  ``Msdn: Mutually semantic distillation network for zero-shot learning,'' in
  \emph{CVPR}, 2022.

\bibitem{xie2019attentive}
G.-S. Xie, L.~Liu, X.~Jin, F.~Zhu, Z.~Zhang, J.~Qin, Y.~Yao, and L.~Shao,
  ``Attentive region embedding network for zero-shot learning,'' in
  \emph{CVPR}, 2019.

\bibitem{jiang2018learning}
H.~Jiang, R.~Wang, S.~Shan, and X.~Chen, ``Learning class prototypes via
  structure alignment for zero-shot recognition,'' in \emph{ECCV}, 2018.

\bibitem{welinder2010caltech}
P.~Welinder, S.~Branson, T.~Mita, C.~Wah, F.~Schroff, S.~Belongie, and
  P.~Perona, ``Caltech-ucsd birds 200. technical report cns-tr-2010-001,''
  \emph{California Institute of Technology}, 2010.

\bibitem{patterson2014sun}
G.~Patterson, C.~Xu, H.~Su, and J.~Hays, ``The sun attribute database: Beyond
  categories for deeper scene understanding,'' \emph{International Journal of
  Computer Vision}, vol. 108, no.~1, pp. 59--81, 2014.

\bibitem{xian2018zero}
Y.~Xian, C.~H. Lampert, B.~Schiele, and Z.~Akata, ``Zero-shot learning—a
  comprehensive evaluation of the good, the bad and the ugly,'' \emph{IEEE
  Transactions on Pattern Analysis and Machine Intelligence}, vol.~41, no.~9,
  pp. 2251--2265, 2018.

\bibitem{pourpanah2020review}
F.~Pourpanah, M.~Abdar, Y.~Luo, X.~Zhou, R.~Wang, C.~P. Lim, and X.-Z. Wang,
  ``A review of generalized zero-shot learning methods,'' \emph{arXiv preprint
  arXiv:2011.08641}, 2020.

\bibitem{li2019leveraging}
J.~Li, M.~Jing, K.~Lu, Z.~Ding, L.~Zhu, and Z.~Huang, ``Leveraging the
  invariant side of generative zero-shot learning,'' in \emph{CVPR}, 2019.

\bibitem{kong2022compactness}
X.~Kong, Z.~Gao, X.~Li, M.~Hong, J.~Liu, C.~Wang, Y.~Xie, and Y.~Qu,
  ``En-compactness: Self-distillation embedding \& contrastive generation for
  generalized zero-shot learning,'' in \emph{CVPR}, 2022.

\bibitem{elhoseiny2021imaginative}
M.~Elhoseiny, D.~Jha, K.~Yi, and I.~Skorokhodov, ``Imaginative walks:
  Generative random walk deviation loss for improved unseen learning
  representation,'' \emph{arXiv preprint arXiv:2104.09757}, 2021.

\bibitem{elhoseiny2021cizsl++}
M.~Elhoseiny, K.~Yi, and M.~Elfeki, ``Cizsl++: Creativity inspired generative
  zero-shot learning,'' \emph{arXiv preprint arXiv:2101.00173}, 2021.

\bibitem{verma2018generalized}
V.~K. Verma, G.~Arora, A.~Mishra, and P.~Rai, ``Generalized zero-shot learning
  via synthesized examples,'' in \emph{CVPR}, 2018.

\bibitem{zhao2017zero}
B.~Zhao, X.~Sun, Y.~Yao, and Y.~Wang, ``Zero-shot learning via
  shared-reconstruction-graph pursuit,'' \emph{arXiv preprint
  arXiv:1711.07302}, 2017.

\bibitem{hu2018correction}
R.~L. Hu, C.~Xiong, and R.~Socher, ``Correction networks: Meta-learning for
  zero-shot learning,'' 2018.

\bibitem{huynh2020fine}
D.~Huynh and E.~Elhamifar, ``Fine-grained generalized zero-shot learning via
  dense attribute-based attention,'' in \emph{CVPR}, 2020.

\bibitem{Yang2021OnIA}
S.~Yang, K.~Wang, L.~Herranz, and J.~van~de Weijer, ``On implicit attribute
  localization for generalized zero-shot learning,'' \emph{IEEE Signal
  Processing Letters}, vol.~28, pp. 872--876, 2021.

\bibitem{Zhang2020APZ}
H.~Zhang, H.~Mao, Y.~Long, W.~Yang, and L.~Shao, ``A probabilistic zero-shot
  learning method via latent nonnegative prototype synthesis of unseen
  classes,'' \emph{IEEE Transactions on Neural Networks and Learning Systems},
  vol.~31, pp. 2361--2375, 2020.

\bibitem{shigeto2015ridge}
Y.~Shigeto, I.~Suzuki, K.~Hara, M.~Shimbo, and Y.~Matsumoto, ``Ridge
  regression, hubness, and zero-shot learning,'' in \emph{ECML-KDD}, 2015.

\bibitem{le2020contrastive}
P.~H. Le-Khac, G.~Healy, and A.~F. Smeaton, ``Contrastive representation
  learning: A framework and review,'' \emph{IEEE Access}, 2020.

\bibitem{khosla2020supervised}
P.~Khosla, P.~Teterwak, C.~Wang, A.~Sarna, Y.~Tian, P.~Isola, A.~Maschinot,
  C.~Liu, and D.~Krishnan, ``Supervised contrastive learning,'' in
  \emph{NeurIPS}, 2020.

\bibitem{han2021contrastive}
Z.~Han, Z.~Fu, S.~Chen, and J.~Yang, ``Contrastive embedding for generalized
  zero-shot learning,'' in \emph{CVPR}, 2021.

\bibitem{vaswani2017attention}
A.~Vaswani, N.~Shazeer, N.~Parmar, J.~Uszkoreit, L.~Jones, A.~N. Gomez,
  {\L}.~Kaiser, and I.~Polosukhin, ``Attention is all you need,'' in
  \emph{NeurIPS}, 2017.

\bibitem{devlin2018bert}
J.~Devlin, M.-W. Chang, K.~Lee, and K.~Toutanova, ``Bert: Pre-training of deep
  bidirectional transformers for language understanding,'' \emph{arXiv preprint
  arXiv:1810.04805}, 2018.

\bibitem{brown2020language}
T.~Brown, B.~Mann, N.~Ryder, M.~Subbiah, J.~D. Kaplan, P.~Dhariwal,
  A.~Neelakantan, P.~Shyam, G.~Sastry, A.~Askell \emph{et~al.}, ``Language
  models are few-shot learners,'' in \emph{NeurIPS}, 2020.

\bibitem{dosovitskiy2020image}
A.~Dosovitskiy, L.~Beyer, A.~Kolesnikov, D.~Weissenborn, X.~Zhai,
  T.~Unterthiner, M.~Dehghani, M.~Minderer, G.~Heigold, S.~Gelly \emph{et~al.},
  ``An image is worth 16x16 words: Transformers for image recognition at
  scale,'' in \emph{ICLR}, 2020.

\bibitem{alamri2021multi}
F.~Alamri and A.~Dutta, ``Multi-head self-attention via vision transformer for
  zero-shot learning,'' \emph{arXiv preprint arXiv:2108.00045}, 2021.

\bibitem{lin2015bilinear}
T.-Y. Lin, A.~RoyChowdhury, and S.~Maji, ``Bilinear cnn models for fine-grained
  visual recognition,'' in \emph{ICCV}, 2015.

\bibitem{guo2018zero}
Y.~Guo, G.~Ding, J.~Han, and S.~Tang, ``Zero-shot learning with attribute
  selection,'' in \emph{Proceedings of the AAAI Conference on Artificial
  Intelligence}, vol.~32, no.~1, 2018.

\bibitem{rahman2018unified}
S.~Rahman, S.~Khan, and F.~Porikli, ``A unified approach for conventional
  zero-shot, generalized zero-shot, and few-shot learning,'' \emph{IEEE
  Transactions on Image Processing}, vol.~27, no.~11, pp. 5652--5667, 2018.

\bibitem{keshari2020generalized}
R.~Keshari, R.~Singh, and M.~Vatsa, ``Generalized zero-shot learning via
  over-complete distribution,'' in \emph{CVPR}, 2020.

\bibitem{chen2021free}
S.~Chen, W.~Wang, B.~Xia, Q.~Peng, X.~You, F.~Zheng, and L.~Shao, ``Free:
  Feature refinement for generalized zero-shot learning,'' in \emph{ICCV},
  2021.

\bibitem{Yu2020EpisodeBasedPG}
Y.~Yu, Z.~Ji, J.~Han, and Z.~Zhang, ``Episode-based prototype generating
  network for zero-shot learning,'' in \emph{CVPR}, 2020.

\bibitem{Yue2021CounterfactualZA}
Z.~Yue, T.~Wang, H.~Zhang, Q.~Sun, and X.~Hua, ``Counterfactual zero-shot and
  open-set visual recognition,'' in \emph{CVPR}, 2021.

\bibitem{Feng2021TransferIF}
L.~Feng and C.~Zhao, ``Transfer increment for generalized zero-shot learning,''
  \emph{IEEE Transactions on Neural Networks and Learning Systems}, vol.~32,
  pp. 2506--2520, 2021.

\bibitem{Min2020DomainAwareVB}
S.~Min, H.~Yao, H.~Xie, C.~Wang, Z.~Zha, and Y.~Zhang, ``Domain-aware visual
  bias eliminating for generalized zero-shot learning,'' in \emph{CVPR}, 2020.

\bibitem{liu2019attribute}
Y.~Liu, J.~Guo, D.~Cai, and X.~He, ``Attribute attention for semantic
  disambiguation in zero-shot learning,'' in \emph{ICCV}, 2019.

\bibitem{Huynh2020FineGrainedGZ}
D.~T. Huynh and E.~Elhamifar, ``Fine-grained generalized zero-shot learning via
  dense attribute-based attention,'' in \emph{CVPR}, 2020.

\bibitem{alwani2022decore}
M.~Alwani, Y.~Wang, and V.~Madhavan, ``Decore: Deep compression with
  reinforcement learning,'' in \emph{Proceedings of the IEEE/CVF Conference on
  Computer Vision and Pattern Recognition}, 2022, pp. 12\,349--12\,359.

\bibitem{lin2017feature}
T.-Y. Lin, P.~Doll{\'a}r, R.~Girshick, K.~He, B.~Hariharan, and S.~Belongie,
  ``Feature pyramid networks for object detection,'' in \emph{Proceedings of
  the IEEE conference on computer vision and pattern recognition}, 2017, pp.
  2117--2125.

\end{thebibliography}

 




\vfill

\end{document}